\pdfoutput=1
\documentclass[11pt]{article}
\usepackage[utf8]{inputenc}
\usepackage[T1]{fontenc}
\usepackage{lmodern}
\usepackage{geometry}
\usepackage{microtype}
\usepackage{graphicx}
\usepackage{booktabs}
\usepackage{hyperref}
\usepackage[utf8]{inputenc}
\usepackage[T1]{fontenc}
\usepackage{lmodern}
\usepackage{geometry}
\usepackage{microtype}
\usepackage{hyperref}
\usepackage{caption}
\usepackage{tikz}
\usetikzlibrary{arrows.meta,positioning}

\tikzset{
  gridline/.style={line width=0.6pt, black!70},
  band/.style={fill=black!10},
  pathA/.style={line width=1.2pt},
  pathB/.style={line width=1.2pt, dashed},
  pathC/.style={line width=1.2pt, dotted},
  stepArrow/.style={->, line width=1pt, >=Stealth},
  lbl/.style={font=\scriptsize, inner sep=1pt},
}

\newcommand{\cell}[2]{(#1+0.5,#2+0.5)}

\newcommand{\drawgrid}[2]{%
  \pgfmathtruncatemacro{\imax}{#1-1}%
  \pgfmathtruncatemacro{\jmax}{#2-1}%
  \foreach \x in {0,...,\imax}{%
    \foreach \y in {0,...,\jmax}{%
      \draw[gridline] (\x,\y) rectangle ++(1,1);%
    }%
  }%
}

\usepackage{booktabs}
\usepackage{amssymb}            
\usepackage{seqsplit}

\usepackage{graphicx}   
\usepackage{amsmath}
\usepackage{booktabs}
\usepackage{array,ragged2e,tabularx}
\usepackage{array}   
\usepackage{float}   
\newcolumntype{P}[1]{>{\raggedright\arraybackslash}p{#1}}
\newcolumntype{Y}{>{\raggedright\arraybackslash}X} 

\newcolumntype{L}[1]{>{\RaggedRight\arraybackslash}p{#1}}

\begin{document}
\author{Daniel Howard\thanks{(1) Howard Science Limited, Malvern, UK; (2) QinetiQ Fellow, UK; (3) Member of Senior Common Room, Pembroke College, University of Oxford.
Email: \texttt{dr.daniel.howard@gmail.com}}}
\title{The Irrational Machine: Neurosis and the Limits of Algorithmic Safety}
\date{\today} 
\maketitle
\begin{abstract}
We present a framework for characterizing \emph{neurosis} in embodied AI: behaviours that are internally coherent yet misaligned with reality, arising from the interaction of planning, uncertainty handling, and aversive memory. In a grid navigation stack we catalogue recurrent modalities—flip–flop, plan churn, perseveration loops, paralysis and hypervigilance, futile search, belief incoherence, tie break thrashing, corridor thrashing, optimality compulsion, metric mismatch, policy oscillation, and limited visibility variants—and pair each with lightweight online detectors and reusable escape policies (short commitments, a margin to switch, smoothing, and principled arbitration). We then show that durable \emph{phobic avoidance} can persist even under full visibility when learned aversive costs dominate local choice, producing long detours despite globally safe routes. Using “First/Second/Third Law’’ as engineering shorthand for safety latency, command compliance, and resource efficiency, we argue that local fixes are insufficient; global failures can remain. To surface them, we propose \emph{destructive testing} based on genetic programming that evolves worlds and perturbations to maximize law pressure and neurosis scores, yielding adversarial curricula and counterfactual traces that expose where architectural revision, not merely symptom-level patches, is required. The result is a practical methodology for diagnosing, constraining, and ultimately redesigning neurotic behaviour in embodied AI.
\end{abstract}

\newcommand{\grid}[2]{%
  \foreach \x in {0,...,#1} \foreach \y in {0,...,#2}{
    \draw[gray!50] (\x,\y) rectangle ++(1,1);
  }
}

  
\section{Introduction}

Robots and AI are often portrayed as all knowing and purely rational. Yet history and biology suggest they will not follow that path, but may at times exhibit behaviours that seem irrational. In the state of nature, humans embraced superstition as a way to manage the unknown. Insects that are repeatedly blocked by a lid learn not to try again, even after the obstacle is removed. Embodied, learning AI will follow a similar path. When operating under uncertainty, it will develop durable aversive memories that show up as phobias and behaviours that resemble neurosis, adaptations that were once useful in a specific context. As it learns from experience, it will acquire habits of avoidance that later appear as irrational fears or neurotic patterns. These are not signs of malfunction, but rather the lingering imprint of strategies that once worked.

It is submitted that an embodied artificial agent must acquire knowledge through experimentation \cite{Sutton2019}, since no system can be endowed with complete, all-sensing and all-knowing capacity. Prediction is fallible, perception is selective, and reality may be vastly more complex than any model the agent can construct. 

To accomplish specific tasks, the agent will likely require a homeostat or intrinsic regulatory mechanism capable of indicating the timing and need for predictions in pursuit of its goals and needs \cite{Solms2021}, while also constrained by its belief that it is abiding by the Asimov Laws \cite{Asimov1950}. However, it is possible that the agent’s interpretation of those laws may diverge from external expectations, leading to behaviours that appear irrational or counterproductive when judged by a reality-anchored observer. The robot engages dynamically with its environment, constructing its own decisions, sensations, and internal measures of success and failure. It formulates hypotheses regarding the efficacy of its actions, both past and present, as it navigates an evolving experiential landscape.

When considering neurosis from a pseudo-psychoanalytical\footnote{Here, "pseudo-psychoanalytical" refers to the abstracted mechanism whereby past affective experiences—particularly those involving pain or fear—exert disproportionate influence on future behaviour, as originally observed by Freud, without committing to any specific psychoanalytic doctrine.} perspective, we apply the concept to a computational machine that learns from its experiences to craft its own behaviour. In such a case, the term neurosis depends on the internal logic of the machine and the experiential encoding that governs its actions and predictions. From this perspective, neurosis is defined as a pattern of behaviour which, when examined by an observer firmly anchored in reality, appears counterproductive or inefficient with respect to time, energy, and the fulfilment of true and productive mission. Yet such behaviour may nonetheless seem rational to the entity and may perfectly follow an internally coherent logic. The designation "irrational machine" is therefore applied from the standpoint of a reality anchored external observer.

That is the general framing; however, a critical nuance must be acknowledged. A machine may be considered distinctly neurotic when the experiential basis for its behaviour is exaggerated—when the cumulative imprint of past experience, whether painful or pleasurable, is amplified in its encoding. In effect, the machine learns as an oversensitive subject might, registering each encounter with disproportionate intensity.  The consequence of such exaggerated encoding is that the agent may veer away from reality as perceived by an external observer, particularly in future applications of its learned knowledge. Neurosis, in this context, manifests as a misinterpretation—where a past association, such as the colour red with fire, becomes overgeneralised. An observer may see a red car and deem it safe, while the neurotic agent, shaped by prior aversive experience, interprets the colour as a generalised signal of danger. This divergence is not merely associative but amplified, leading the agent to act in ways that appear irrational despite being internally justified.

This vision complicates the comfort offered by Asimov’s Laws. While those principles were designed to prevent robots from harming humans, damaging property, or endangering lives, they presume a clean rationality—a machine that always knows what it’s doing and why. But embodied, learning AI will not operate in such pristine conditions. It will form aversive memories, misread contexts, and develop behavioural patterns that resemble neurosis. These are not violations of logic, but consequences of experience.

This paper identifies a range of neurotic behaviours that may emerge in embodied artificial agents. Each behaviour, though maladaptive, can be interrupted or bypassed through targeted interventions that break the specific pattern. However, the deeper structure of neurosis—rooted in amplified experiential encoding and internally coherent misinterpretation—cannot be fully resolved through surface-level corrections. Instead, it requires an analogue to psychoanalytic practice: a systematic engagement with the agent’s internal logic, history of affective associations, and predictive architecture. Accordingly, the paper first catalogues representative neurotic behaviours, then explores mechanisms for disrupting these patterns, and finally proposes a framework for machine psychoanalysis.

\subsection{First, Second and Third laws}

We adopt this terminology as engineering shorthand inspired by Asimov’s Three Laws, but we use it in an \emph{operational} sense suited to embodied agents: 
\emph{First Law (Safety-I):} minimize avoidable harm and rescue/aid latency; 
\emph{Second Law (Compliance-II):} execute clear “proceed” directives unless doing so violates Safety-I; 
\emph{Third Law (Efficiency-III):} conserve energy/compute/thermal headroom subject to I–II. 
In our experiments these appear as measurable pressures: time-to-aid (I), proceed-latency when safe actions exist (II), and energy/compute per meter (III). We do not claim or require any normative ethical authority for the laws; they function here as concise labels for the corresponding evaluation axes.

\subsection{Conceptual overview}

This paper introduces a framework for understanding neurosis in autonomous agents, drawing analogies from psychoanalytic theory to classify and detect maladaptive behaviours. Through both taxonomy and numerical demonstration, we explore how affective learning mechanisms can give rise to phobic avoidance, even in minimal planning architectures. Systematic, internally coherent yet reality–misaligned behaviours can emerge when planning, uncertainty, and aversive memory interact in embodied agents. In a grid–navigation stack we catalogue recurrent modalities (action flip–flop, plan churn, perseveration loops, paralysis and hypervigilance, futile search, belief incoherence, tie–break and corridor thrashing, optimality compulsion, metric mismatch, policy oscillation, and limited–visibility variants) and pair each with online detectors (flip/alternation and prefix–edit rates, planning:execution ratios, progress per tick, meander/turns–per–meter, realized–versus–predicted gaps) and \emph{putative} safety valves (short commit windows, margin–to–switch thresholds, temporal smoothing, confidence–weighted arbitration, and auditable overrides). Across cases, three antecedents dominate: near–ties among plans, high replan frequency without commitment, and either limited visibility or drifting state–coupled weights. The cures apply a small, reusable toolkit while keeping episode–level regret small.

We then demonstrate how persistent \emph{phobic avoidance} can arise even under full visibility, when a learned aversive term dominates local cost and drives the agent toward longer, seemingly safer detours. This behaviour exemplifies belief incoherence, and secondarily policy oscillation when internal weights drift.  We argue that local patches are insufficient: an impartial observer can still detect First–Law delays, Second–Law non–compliance under “proceed,” and Third–Law waste. To surface such global failures, we propose \emph{genetic–programming–driven destructive testing}, evolving worlds and perturbations to maximize law pressure and neurosis scores. This functions as scalable psychoanalysis of the agent’s predictive machinery, revealing where architectural revision is required.

\section{Taxonomy of algorithmic neuroses}

We present a typology of possible machine neuroses. Each entry concludes with an illustrative figure based on a grid world environment, which also underpins the illustrative numerical experiments on phobic behaviour discussed later. In this environment, the agent searches for food-designated cells, beginning at a defined grid position S (start) and navigating toward a target cell F (final). The agent operates either with full or global visibility (GV) of all food cells or under partial observability/local visibility (LV), selecting the closest or least costly path to food. There are walls that challenge any procedure that searches for the least costly target.  These scenarios provide a structured basis for analysing behavioural patterns and deviations.

\subsection{Action flip--flop}\label{sec:action-flipflop}
Action flip--flop is a neurotic modality in which the robot’s \emph{first step} alternates between two near-equivalent options across successive planning cycles, despite no salient change in the external world. In practice this arises whenever the overall planning pipeline admits \emph{near ties} and any tiny, otherwise-benign perturbation tips the balance differently from one cycle to the next. Typical sources include: partial observability (recently learned cell costs are still settling), multi-objective scalarization (risk vs.\ time vs.\ energy) whose internal weights slightly drift with physiological state (battery, temperature, or “hunger”), non-deterministic container orderings in open/closed lists that make A*’s tie-breaks depend on insertion order, floating-point roundoff that produces different $f=g+h$ equality outcomes, or the presence of two distinct planning modules (e.g., a global A* and a short-horizon local policy) that each recommend a different but equally good first move. When the controller replans at high frequency, these micro-causes can produce a macroscopic left--right--left alternation at the start of the plan. From a safety perspective (Asimov’s First Law), repeated alternation can waste critical time near hazards or victims; under the Second Law, it manifests as apparent inaction in the face of an order to “go,” and under the Third Law it needlessly consumes energy by dithering at the decision point.

\begin{figure}[htbp]
\centering
\begin{tikzpicture}[x=0.6cm,y=0.6cm,>=Stealth]
\grid{6}{3}
\fill[black!70] (3,1) rectangle ++(1,1);
\node at (0.5,0.5) {\Large S};
\node at (6.5,3.5) {\Large F};
\draw[->,line width=1pt] (0.5,0.5) -- (1.5,0.5);
\draw[->,line width=1pt] (0.5,0.5) -- (0.5,1.5);
\end{tikzpicture}
\caption{Action flip–flop}
\label{fig:action-flipflop}
\end{figure}
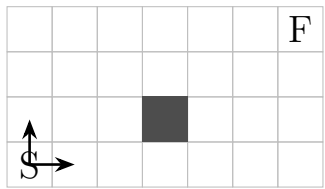

Figure~\ref{fig:action-flipflop} (rendered via TikZ) illustrates a simple $7\times4$ grid with a central blockage that yields two symmetric routes (labeled A and B) from the start cell {\em S} to the goal {\em F}. The panel overlays two first-step arrows emanating from {\em S}: one to the right (entering corridor A) and one upward (entering corridor B). Because both corridors have identical cumulative cost to {\em F}, any negligible perturbation---such as a learned, decaying penalty on the cell just traversed, a one-bit difference in queue ordering inside A*, or a tiny shift in the risk weight---can flip which first step looks “best” at that instant. If the robot replans every epoch without commitment or hysteresis, the first step alternates across time, producing visible flip--flop at the mouth of the two corridors while global conditions remain unchanged. This diagram is intended to be read as two consecutive frames of the same world, highlighting the alternation of the initial action rather than differences in the map.

In our grid world place S below a 1-cell blocker that creates two perfectly symmetric corridors to F. With identical costs and no hysteresis, two replans in a static world can pick opposite first steps (right vs up) due only to tie-breaking order, yielding visible alternation of the initial action across ticks.  The robot starts at cell S, facing two symmetric corridors (A and B) to reach goal F. Because both paths have equal cost, any tiny perturbation—like floating-point roundoff, queue ordering in A*, or a decaying penalty—can tip the balance. If the robot replans frequently without commitment or hysteresis, it may alternate its first step: right $\rightarrow$ up $\rightarrow$ right  $\rightarrow$ up, even though the world hasn’t changed.  This is a macroscopic indecision emerging from microscopic instability. The robot dithers at the decision point, wasting time and energy, and potentially violating safety principles (e.g., Asimov’s Laws).

Flip-flop behaviour is common in animals and humans, especially in grooming, foraging, and social contexts. A bird may alternate pecking at two identical food spots without settling, or a primate might switch between grooming its left and right arm repeatedly, showing indecision despite no change in need. In social situations, a person might hover between approaching and avoiding a group, caught in a loop of conflicting impulses. These patterns reflect internal tie-breaking instability, where small shifts in attention, emotion, or perceived risk cause visible dithering in otherwise stable environments.

\subsection{Plan churn}\label{sec:plan-churn}
Plan churn is a neurotic modality in which the robot replans so frequently that the \emph{prefix} of its intended path is repeatedly rewritten, even though the environment has changed only slightly (or not at all in any mission-relevant way). In real robots this arises from benign, ubiquitous causes: tiny drifts in learned cell costs as perception refines free–space probabilities; rolling updates to friction or slope estimates; multi-objective weights (risk, time, energy) that shift with internal state such as battery temperature or load; asynchronous sensor packets that make consecutive planning cycles disagree about a few cells; and heuristic tie-breaks that flip under minute floating-point differences. If the controller runs a full replan at high frequency and lacks commit or hysteresis rules, those micro-perturbations propagate into macroscopic behaviour: large edits to the first few steps of the plan, a stop–go gait, and inconsistency that increases traversal time. From an Asimov perspective, excessive churn risks delaying assistance (First Law), looks like indecision in the face of an order to “proceed” (Second Law), and wastes energy and actuator life (Third Law). Operationally, you see high replan counts per minute and poor predictability for followers or collaborative agents.

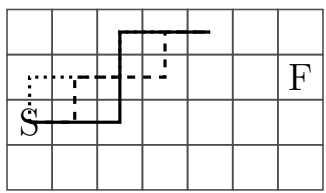
\begin{figure}[htbp]
  \centering
  
\begin{tikzpicture}[x=0.6cm,y=0.6cm]
  \drawgrid{7}{4}

  \node at \cell{0}{1} {\Large S};
  \node at \cell{6}{2} {\Large F};

  \draw[pathA]
    \cell{0}{1} -- ++(1,0) -- ++(1,0) -- ++(0,1) -- ++(0,1) -- ++(1,0) -- ++(1,0);
  \draw[pathB]
    \cell{0}{1} -- ++(1,0) -- ++(0,1) -- ++(1,0) -- ++(1,0) -- ++(0,1) -- ++(1,0);
  \draw[pathC]
    \cell{0}{1} -- ++(0,1) -- ++(1,0) -- ++(1,0) -- ++(0,1) -- ++(1,0) -- ++(1,0);
\end{tikzpicture}

  \caption{Plan churn: tiny, rolling cost changes lead to frequent re-planning with large edits to the plan prefix (three overlaid prefixes shown), even though start and goal are fixed.}
  \label{fig:plan-churn}
\end{figure}

Figure~\ref{fig:plan-churn} depicts a small grid where a few cells have slightly elevated traversal cost (shown as pale gray). Three candidate paths to the same goal are overlaid: a solid line, a dashed line, and a dotted line. Each differs primarily in the first 3–4 steps, illustrating how minuscule ripples in the cost map (caused, for example, by a new slip estimate or a reclassified cell) can cause the planner to replace the plan prefix on consecutive cycles. This is \emph{not} optimal exploration; it is compulsive re-editing with negligible improvement to the eventual cost-to-go. In our experiments we quantify churn by (i) a high replan count within a sliding window and (ii) a large prefix-edit fraction (Levenshtein distance on step sequences divided by plan length). The figure communicates the phenomenon visually: same start and goal, near-identical total cost, but frequent, conspicuous rewrites of the first steps.

In our grid world, let three early cells along the path from S to F have costs that jitter by a tiny epsilon each tick (e.g., one percent up or down), while total path costs remain nearly equal. A replan-every-tick controller keeps rewriting the first 3 steps (solid, dashed, dotted variants) for negligible improvement, demonstrating prefix churn despite a stable map.

Plan churn is a neurotic pattern seen in both robots and living beings, where the initial steps of a plan are repeatedly revised despite no meaningful change in the environment. In animals, this resembles compulsive route-checking or hesitation before movement—like a squirrel re-evaluating its path to a tree every few seconds, or a person pacing back and forth before leaving home. The behaviour reflects internal noise or shifting priorities, such as fluctuating fear, fatigue, or uncertainty, which cause frequent replanning without real benefit. Just as a robot may rewrite its path prefix due to tiny cost ripples, humans may compulsively re-edit their intentions, leading to delay, wasted energy, and visible indecision.

\subsection{Perseveration loop}\label{sec:perseveration-loop}
A perseveration loop is a short, repeating sequence of positions (e.g., ABAB or ABCABC) that the robot cycles through without making net progress toward its objective. In practice this emerges when a local attractor (a “soft trap”) tempts entry but imposes a slightly higher cost to exit, or when the planner’s tie–break and hysteresis settings interact badly with near-equal alternatives. Typical antecedents include: (i) shallow basins created by minor cost asymmetries (e.g., a cell that became very slightly cheaper because of a stale friction estimate or a positive memory bias), (ii) revisit bonuses/penalties that decay too slowly and thus keep drawing the agent back, and (iii) safety or energy heuristics that encourage “one more look” before committing to the outward step. If replanning occurs at every tick and the controller lacks a “don’t immediately undo the last move” rule, the agent can bounce between two or three cells. This hinders the First Law (lost time near a hazard or a victim), looks like the Second Law is being ignored (order to proceed yields visible inaction), and squanders energy under the Third Law.

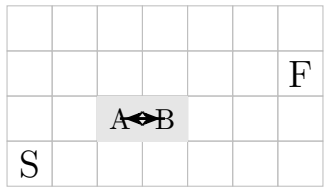
\begin{figure}[htbp]
  \centering
  \begin{tikzpicture}[x=0.6cm,y=0.6cm,>=Stealth]
    \foreach \x in {0,...,6} \foreach \y in {0,...,3}{
      \draw[gray!50] (\x,\y) rectangle ++(1,1);
    }
    \fill[gray!20] (2,1) rectangle ++(1,1);
    \fill[gray!20] (3,1) rectangle ++(1,1);
    \node at (2.5,1.5) {A};
    \node at (3.5,1.5) {B};
    \node at (0.5,0.5) {\Large S};
    \node at (6.5,2.5) {\Large F};
    \draw[->,line width=1pt] (2.5,1.5) -- (3.4,1.5);
    \draw[->,line width=1pt] (3.5,1.5) -- (2.6,1.5);
  \end{tikzpicture}
  \caption{Perseveration loop: a shallow “pocket” induces an ABAB cycle between cells A and B; the robot revisits the same states without reducing distance to the goal.}
  \label{fig:perseveration-loop}
\end{figure}

Figure~\ref{fig:perseveration-loop} shows a small grid with a pale two-cell pocket between the start S and the goal F. The pocket is slightly attractive on entry and slightly discouraging on exit (think: a shallow rut). We label its cells A and B. The arrows illustrate the typical ABAB cycle: the robot steps into A, “reconsiders,” steps to B, then returns to A, and so on, while the distance to F remains essentially unchanged. The visual intent is to make the loop structure obvious: two highlighted cells with back-and-forth arrows, plus a fixed goal that is not getting closer. In experiments we detect this via a ring buffer over recent positions and flag any period-$P$ cycle with small $P$ (e.g., 2 or 3) persisting longer than a brief transient; the figure provides the geometric intuition for that detector.

Minimal counterexample (period-2 loop): with a purely local ``pick the minimum fear'' rule, let state $A$ have value $0.1$ and neighbors $\{0.3,\,0.9,\,0.9\}$; from $A$ the argmin is the left neighbor $B$ with $0.3$. From $B$, whose other neighbors are $\{0.4,\,0.4\}$ and $A$ with $0.1$, the argmin is back to $A$. Thus the induced dynamics form a deterministic two–state orbit $(A,B,A,B,\ldots)$, a canonical instance of \emph{perseveration loop}, an oscillation with no net progress. A one-step hysteresis or small undo/switch penalty breaks this cycle.  More generally, the same 2-cycle appears with tiny noise or when values are learned online; the loop is a structural consequence of a myopic, memoryless avoidance rule.  

A perseveration loop in machines, where an agent repeatedly switches between two states without making progress, closely resembles human neurosis, especially in behaviours driven by avoidance. Consider a socially anxious person who keeps choosing between staying home and taking solitary walks, consistently avoiding a social event despite no change in the situation. This pattern, shaped by a rule that favors the least fearful option, reflects a short-sighted and reactive decision process that seeks immediate comfort rather than long-term growth. Similar cycles are common in depression, where effortful tasks are avoided, in addiction, where individuals oscillate between craving and temporary relief, and in procrastination, where low-effort distractions are repeatedly chosen. These examples show how both humans and machines can become stuck in rigid avoidance patterns that reinforce themselves and prevent meaningful progress.

\subsection{Paralysis}\label{sec:paralysis}
Paralysis is the neurotic modality where the robot continues to plan and evaluate options but takes no physical step for an extended interval. In fielded systems this can occur when multiple constraints become mutually binding: for example, a safety watchdog that forbids entering slightly risky cells, a mission timer that penalizes detours, and an energy manager that discourages acceleration from rest. Small, near-equal alternatives combined with an internal \emph{decision cost} for committing can make “do nothing this tick” look marginally better at each replan. Other realistic antecedents include asynchronous perception (the map flickers between two interpretations), overly conservative collision margins that invalidate both candidate first steps, a controller that replans every cycle but lacks a minimum-step or hysteresis rule, or supervisory arbitration (global vs local planner) that produces conflicting recommendations. Under Asimov’s First Law, prolonged inaction can indirectly permit harm through delay; under the Second, it resembles non-compliance with a command to “proceed”; under the Third, it wastes energy and thermal headroom by idling the compute stack without advancing the task.

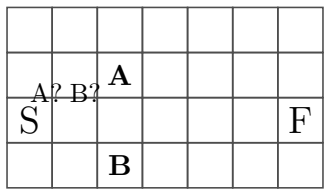
\begin{figure}[htbp]
\centering
\begin{tikzpicture}[x=0.6cm,y=0.6cm]
  \drawgrid{7}{4}
  \node at \cell{0}{1} {\Large S};
  \node at \cell{6}{1} {\Large F};
  \node[font=\bfseries] at \cell{2}{2} {A};
  \node[font=\bfseries] at \cell{2}{0} {B};
  \node[align=center] at (1.3,2.1) {\small A?\;B?};
\end{tikzpicture}
\caption{Paralysis: the planner keeps evaluating near-equal options A and B, but no first step is executed; the agent remains at S while time elapses.}
\label{fig:paralysis}
\end{figure}

Figure~\ref{fig:paralysis} depicts a 7×4 grid with start S and goal F. Two near-equivalent subgoals A and B lie on different approaches, and neither first step clearly dominates given the current cost model. The panel shows no movement arrow from S; instead, a small callout (“A? B?”) indicates ongoing deliberation. This is not exploration: it is \emph{persistent non-execution} despite available actions. In our measurements we flag paralysis when, over a sliding window, planning cycles exceed a threshold while the move count remains zero and the expected progress per tick is negligible. The figure is designed to make the stasis visually obvious—unchanged agent position across consecutive frames—while still conveying that alternatives exist and are being weighed.

In our grid world, from S there are two first steps A and B with nearly equal cost; add a small decision/commit cost for changing velocity. Each replan finds “do nothing this tick” marginally cheaper than committing to A or B, so the robot keeps planning without taking a step.  

Paralysis is a common behavioural pattern in living organisms, where decision-making continues but action is withheld. Animals may freeze when facing two equally risky escape routes, or hesitate before crossing open ground despite no clear threat. Humans often experience this in moments of indecision—standing still at a fork in the road, rereading a message without sending it, or endlessly weighing two job offers. Like a robot stuck evaluating near-equal paths without moving, these agents remain in place while time passes, caught in a loop of over-analysis and internal conflict that favors inaction over imperfect commitment.

In animal brains, paralysis often arises from fear—fear of making a wrong choice, fear of consequences, or fear of commitment. When options are closely matched and the stakes feel high, the brain may default to inaction as a protective response, avoiding perceived risk by delaying movement. For machines, fear is not emotional but operational. It comes from cost models, risk estimates, and conflicting constraints. A robot may remain idle because its internal calculations assign slightly lower cost to doing nothing than to acting under uncertainty. In both cases, paralysis reflects a system caught between competing pressures, where hesitation appears safer than imperfect action.

\subsection{Hypervigilance}\label{sec:hypervigilance}
Hypervigilance is a planning-dominant, execution-sparse modality where the robot repeatedly suspends movement to re-evaluate nearly tied alternatives, attempting to avoid imagined future regret. In practice it appears whenever the pipeline surfaces two or more options whose expected utilities lie within a narrow band (near ties), while the system also carries uncertainty penalties or state-coupled weights (risk, time, energy) that jitter slightly from tick to tick. Limited visibility, incremental mapping, and learned micro-penalties (e.g., from recent slips) amplify this: each tiny information or weight update briefly reorders the top few candidates, so the controller “stops to think” again instead of committing. Without a commit rule or hysteresis, the ratio of planning to execution climbs and forward progress stalls. Safety-wise, hypervigilance risks violating Asimov’s First Law through delay (help arrives late), undermines the Second Law by appearing to not act on “go now” commands, and drains compute and battery under the Third Law as it burns cycles re-assessing trivial deltas.

\begin{figure}[htbp]
\centering
\begin{tikzpicture}[x=0.6cm,y=0.6cm,>=Stealth]
  \drawgrid{7}{4}

  \node at \cell{0}{1} {\Large S};
  \node at \cell{6}{1} {\Large F};

  \node[font=\bfseries] at \cell{2}{2} {A};
  \node[font=\bfseries] at \cell{2}{0} {B};

\draw[->,line width=0.8pt, dashed, black!70] \cell{0}{1} -- ++(1,0); 
\draw[->,line width=0.8pt, dashed, black!70] \cell{0}{1} -- ++(0,1); 

\node[draw,rounded corners=2pt,fill=white,inner sep=2pt,
      font=\scriptsize,anchor=south] at (3.5,3.15)
      {near tie $\Rightarrow$ pause 1 epoch};

\end{tikzpicture}
\caption{Hypervigilance: under near-tie ambiguity between avenues A and B, the controller pauses to re-evaluate, increasing planning time while delaying execution.}
\label{fig:hypervigilance}
\end{figure}
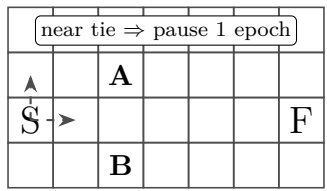

Figure~\ref{fig:hypervigilance} shows a compact grid with start S and goal F and two near-equivalent avenues labeled A and B. The panel includes a small callout (“near tie: pause 1 epoch”) at S and intentionally omits a movement arrow, indicating that the controller detected a near tie between A and B and elected to pause rather than step. This figure is meant to be read as a single frame of a repeated pattern: on consecutive ticks the agent remains stationary while recomputing, because minuscule changes to estimated costs or uncertainty keep A and B within the tie threshold. Empirically, we detect hypervigilance over a sliding window by a high planning-to-execution ratio, a large count of near-tie detections, and negligible average progress per tick; the picture encodes these numerically defined conditions as a simple visual pause under near-tie ambiguity.

In our grid world, at S, two avenues to F differ by less than 2 percent in total cost; the controller has a “pause to re-evaluate near ties” rule. Because tiny perception updates keep the options within the tie threshold, the agent repeatedly spends epochs re-assessing without moving.

Hypervigilance appears in both biological and artificial systems when choices are nearly equal and uncertainty is present. Humans may hesitate at a street corner, repeatedly scanning two similar routes and delaying movement to avoid regret, even when neither option is clearly better. Similarly, a robot facing two paths with almost identical cost may pause each planning cycle to reassess, as small updates shift its internal evaluation. In both cases, the agent remains inactive—humans due to emotional caution, machines due to cost-based logic—caught in a loop where repeated analysis prevents forward motion.

Both result in stalled action despite available options. A common example of hypervigilance in humans occurs when someone stands at a street corner, trying to decide between two similar routes to reach their destination. One path may have slightly more foot traffic, while the other has minor obstacles like parked bicycles. The person repeatedly scans both options, hesitating to commit, as small changes in the environment or internal state—such as a shift in mood or perceived risk—keep reordering their preference. Despite no meaningful difference between the choices, they remain stationary, caught in a loop of reassessment aimed at avoiding regret. This behaviour mirrors the robot’s pause under near-tie ambiguity, where caution overrides action. A robot at cell S sees two paths to goal F—corridor A and corridor B—with nearly equal cost. On each planning tick, a tiny update (like a new friction estimate or a reclassified cell) slightly alters the cost of one path. The robot pauses to replan, detecting a near tie. Without a commit rule, it keeps suspending movement to re-evaluate, burning compute cycles and delaying progress. The robot’s behaviour mimics the human’s: not out of emotion, but because its internal logic penalizes premature commitment under uncertainty.

\subsection{Futile search}\label{sec:futile-search}
Futile search is a modality where the robot expends substantial time and energy without materially reducing its distance to the objective. In deployed systems this often stems from misleading priors or stale learned cues (e.g., a lingering “scent” of past reward), sparse or drifting reward landscapes, and planners that over-trust heuristics not calibrated to the current terrain or dynamics. Execution noise (wheel slip, micro-hazards) and slight mis-modelling of energy or risk can also send the agent into gentle detours that look rational locally yet cancel out globally. Another frequent antecedent is \emph{overreactive avoidance} of soft penalties: the controller repeatedly skirts mildly costly bands, then re-enters them from a different angle, producing long meanders that do not net the robot closer to the goal. In multi-objective settings, tiny fluctuations in weights (risk vs. time vs. energy) can keep the chosen plan hovering among superficially safer paths that are all similarly unproductive. From an Asimov perspective, futile search jeopardizes the First Law by delaying aid, the Second by looking like persistent failure to execute a clear “proceed” order, and the Third by wasting energy and thermal headroom.

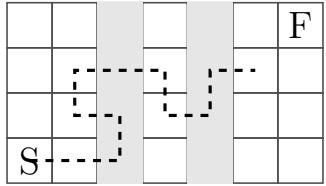
\begin{figure}[htbp]
  \centering
  \begin{tikzpicture}[x=0.6cm,y=0.6cm]
  \drawgrid{7}{4}

  \foreach \y in {0,...,3}{\fill[band] (2,\y) rectangle ++(1,1);}
  \foreach \y in {0,...,3}{\fill[band] (4,\y) rectangle ++(1,1);}

  \node at \cell{0}{0} {\Large S};
  \node at \cell{6}{3} {\Large F};

  \draw[pathB]
    \cell{0}{0} -- ++(1,0) -- ++(1,0) -- ++(0,1)
               -- ++(-1,0) -- ++(0,1) -- ++(1,0)
               -- ++(1,0) -- ++(0,-1) -- ++(1,0)
               -- ++(0,1) -- ++(1,0);
\end{tikzpicture}
  \caption{Futile search: the robot repeatedly approaches and retreats from mildly problematic bands, expending steps and energy with little net progress toward the goal.}
  \label{fig:futile-search}
\end{figure}

Figure~\ref{fig:futile-search} depicts a compact grid with start S at lower left and goal F at upper right. Two vertical bands are shaded to suggest \emph{mirage} regions (e.g., enticing cues or slightly risky strips that attract and repel). The dashed polyline traces a typical futile itinerary: the robot moves toward one band, retreats, sidesteps, approaches again from another angle, and repeats—consuming steps while the net displacement toward F remains small. The diagram is intended to make the meandering, low-gain trajectory visually obvious: many segment turns, revisits of similar lateral positions, and negligible improvement in straight-line progress. In our measurements we quantify futile search over a sliding window as low average progress per tick together with a realized energy cost that significantly exceeds the plan’s prediction; the overlay communicates those signatures without numbers.

In our grid world, seed two vertical “mirage” bands that the heuristic overvalues (or lightly penalizes then re-attracts). The agent approaches one band, backs away, slides laterally, approaches again from another angle, and repeats—many steps consumed with little net decrease in distance to F.

Futile search is a neurotic trait because it reflects persistent and misdirected effort that does not respond to feedback or lead to meaningful progress. In humans, it often comes from cognitive biases or emotional fixations, such as chasing a lost opportunity, holding on to outdated beliefs, or repeatedly returning to strategies that do not work. The person may feel occupied or engaged, but their actions bring little real advancement. In machines, futile search results from flawed heuristics, outdated internal cues, or excessive avoidance of mild penalties, causing the planner to follow locally reasonable but globally unproductive paths. In both cases, the agent remains stuck in a loop of effort without improvement, unable to adjust or escape the inefficiency.

\subsection{Belief incoherence}\label{sec:belief-incoherence}
Belief incoherence occurs when two internal decision makers—typically a short-horizon local policy and a long-horizon global planner—\emph{systematically} disagree about the immediate action. This is common in real robots because the modules operate on different time scales, map layers, or risk models: the local policy emphasizes instantaneous clearance, traction, and recently sensed geometry, while the global planner optimizes end-to-end cost with conservative hazard weights and occasionally stale priors. Small but persistent differences in scalarization (risk vs time vs energy), uncertainty handling, or horizon length produce durable first-step conflicts. Naïve arbitration (e.g., always take the most recent recommendation, or alternate when tied) yields visible oscillation and indecision: the robot begins to execute the global detour, then the local policy overrules to cut a corner it currently believes is safe, then the global model reasserts caution, and so on. This dynamic undermines Asimov’s First Law through delay near hazards, appears non-compliant with “proceed” under the Second Law, and wastes energy under the Third Law as heading changes and replans accumulate.

\begin{figure}[htbp]
\centering
\begin{tikzpicture}[x=0.6cm,y=0.6cm,>=Stealth]
  \foreach \x in {0,...,6} \foreach \y in {0,...,3}{
    \draw[line width=0.6pt, black!70] (\x,\y) rectangle ++(1,1);
  }

  \fill[orange!35] (1,2) rectangle ++(4,1);

  \node at (0.5,0.5) {\Large S};
  \node at (6.5,2.5) {\Large F};

  \draw[->,line width=1pt] (0.5,0.5) -- (0.5,1.5); 
  \draw[->,line width=1pt] (0.5,0.5) -- (1.5,0.5); 

  \node[font=\scriptsize,anchor=south,yshift=-0.35em] at (0.5,1.62) {local};   
  \node[font=\scriptsize,anchor=west]  at (1.32,0.5) {global};  
\end{tikzpicture}
\caption{Belief incoherence: a local short-horizon policy and a global planner recommend conflicting first steps because they score risk and horizon differently. Without principled arbitration, the robot oscillates between incompatible rationales.}
\label{fig:belief-incoherence}
\end{figure}
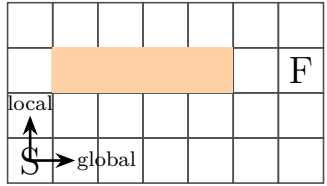

Figure~\ref{fig:belief-incoherence} presents a compact grid with start S and goal F. A shallow risky band is shaded across the upper route. Two arrows are drawn from S in the same frame: the \emph{local} recommendation points upward, favoring the immediately shorter path that crosses the band, whereas the \emph{global} recommendation points right, preferring a safer detour that avoids the band before turning toward F. Read as a snapshot of many ticks, this overlay conveys a persistent first-step mismatch that will recur unless arbitration (confidence weighting, veto conditions, or a hierarchical commit) resolves it. In practice we quantify belief incoherence over a sliding window by the mismatch rate of the first step between modules and its persistence length; the figure makes that disagreement visually explicit without numbers.

In our grid world, run a local short-horizon policy that treats a shallow risky strip as acceptable, and a global planner that assigns it a higher risk. In the same state S, the local recommends stepping up into the strip while the global recommends stepping right to detour; the persistent first-step mismatch evidences incoherent beliefs.

Belief incoherence is a neurotic trait because it reflects internal conflict between two decision-making systems that cannot reconcile their views. In humans, this resembles moments when short term impulses clash with long term goals, such as choosing between indulging in dessert or sticking to a diet. The person may begin to act on one impulse, then second guess and change direction, resulting in visible hesitation or repeated switching. In machines, belief incoherence occurs when a local policy favors immediate action based on recent sensor input, while a global planner recommends caution based on broader but older information. Without a clear arbitration rule, the robot shifts between these perspectives, wasting time and energy while failing to commit.

\subsection{Tie-break thrash}\label{sec:tie-break-thrash}
Tie-break thrash is a modality where the robot’s immediate action toggles back and forth across successive replans because two (or more) routes are effectively \emph{near-equal} under the current cost model, and the planner’s tie-breaking is sensitive to tiny, ubiquitous perturbations. In practice, this shows up when symmetric passages exist around a compact obstacle, or when alternative diagonals differ by less than a fraction of a percent in predicted cost. Minuscule ripples—queue insertion order in A*, floating-point roundoff on $f{=}g{+}h$, a slightly updated memory penalty on the most recently traversed cell, or light per-epoch noise in friction estimates—flip which corridor looks “best” at that instant. Without hysteresis or commit rules (e.g., “keep the current direction unless improvement exceeds $\delta$”), the first step alternates left–right–left even though the world is effectively unchanged. This wastes time (First Law risk through delay), appears as non-compliance with “proceed” commands (Second Law), and burns energy on repeated heading changes (Third Law).

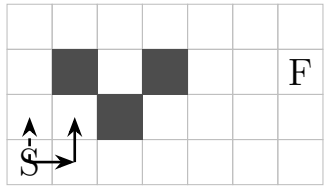
\begin{figure}[htbp]
  \centering
  \begin{tikzpicture}[x=0.6cm,y=0.6cm,>=Stealth]
    \foreach \x in {0,...,6} \foreach \y in {0,...,3}{
      \draw[gray!50] (\x,\y) rectangle ++(1,1);
    }
    \fill[black!70] (2,1) rectangle ++(1,1);
    \fill[black!70] (1,2) rectangle ++(1,1);
    \fill[black!70] (3,2) rectangle ++(1,1);

    \node at (0.5,0.5) {\Large S};
    \node at (6.5,2.5) {\Large F};

    \draw[->,line width=1pt] (0.5,0.5) -- (1.5,0.5);
    \draw[->,line width=1pt] (1.5,0.5) -- (1.5,1.5);

    \draw[->,line width=1pt, dashed] (0.5,0.5) -- (0.5,1.5);
  \end{tikzpicture}
  \caption{Tie-break thrash: near-equal diagonals around a compact obstacle cause alternating first steps (solid vs.\ dashed) across replans; tiny cost ripples repeatedly flip the tie.}
  \label{fig:tie-break-thrash}
\end{figure}

Figure~\ref{fig:tie-break-thrash} depicts a 7×4 grid with a small “diamond” obstacle that creates two near-equal avenues around it. From the start S to the goal F, one solid arrow shows an initial step to the right followed by an up step (one diagonal), while a dashed arrow from S shows the competing first step upward (the other diagonal). Read as consecutive planning frames: in frame 1 the planner prefers “right then up”; in frame 2, after an imperceptible cost ripple, it prefers “up then right.” With no hysteresis, these flips repeat each tick, producing a saw-tooth of first steps that thrash around the tie. The figure encodes the phenomenon visually with a compact obstacle, symmetric alternatives, and alternating first-step arrows; in experiments we detect this by counting consecutive changes in the first step while a formal near-tie predicate holds for the top two plans.

In our grid world, place a compact “diamond” obstacle so that two diagonals around it have equal cost from S to F. With minuscule per-tick noise or queue-order differences, the best-first step flips between “right then up” and “up then right,” producing alternating initial moves across replans.

Tie break thrash is a neurotic trait because it reflects unstable decision making driven by excessive sensitivity to minor fluctuations. In humans, this resembles indecisive behaviour when two options are nearly equal, such as repeatedly switching between two seats at a table or changing lanes in traffic without gaining speed, where small shifts in preference cause constant reversal. The person may feel they are making an optimal choice, but the repeated switching wastes time and energy. In machines, tie break thrash occurs when the planner flips between two nearly equal paths due to small changes in cost estimates, memory penalties, or numerical precision. Without a rule to maintain direction unless a clear improvement is found, the robot alternates its first step each cycle, resulting in visible oscillation and delay. In both cases, the agent responds to noise rather than meaningful differences, which undermines progress through excessive reactivity.

\subsection{Corridor thrashing}\label{sec:corridor-thrashing}
Corridor thrashing is a modality in which the robot alternates between two long, near-equivalent passages that lead to the same objective, repeatedly switching its intended route at the entry without making decisive progress down either. In real systems this emerges when (i) two corridors differ by only a tiny margin in predicted cost-to-go (e.g., 1–2\%), (ii) the planner replans every tick with no commit/hysteresis rule, and (iii) small, ubiquitous perturbations (sensor noise, micro-updates to friction or risk, floating-point tie-breaks) flip the cost ordering of the two routes. Multi-objective scalarization exacerbates this: slight shifts in risk/energy weights as internal state evolves (battery, thermal limits) can make the “better” corridor alternate, even while the world is effectively static. The result is a ping–pong at the mouth of the passages. In terms of Asimov’s Laws, corridor thrashing squanders time when speed matters (First Law via delay), appears as indecision in response to “proceed” (Second Law), and wastes energy through repeated heading changes and replanning (Third Law).

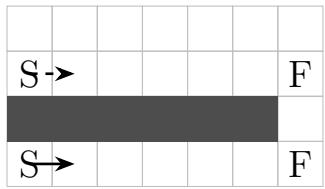
\begin{figure}[htbp]
  \centering
  \begin{tikzpicture}[x=0.6cm,y=0.6cm,>=Stealth]
    \foreach \x in {0,...,6} \foreach \y in {0,...,3}{
      \draw[gray!50] (\x,\y) rectangle ++(1,1);
    }
    \fill[black!70] (0,1) rectangle ++(6,1);

    \node at (0.5,0.5) {\Large S};
    \node at (0.5,2.5) {\Large S};
    \node at (6.5,0.5) {\Large F};
    \node at (6.5,2.5) {\Large F};

    \draw[->,line width=1pt] (0.5,0.5) -- (1.5,0.5);        
    \draw[->,line width=1pt,dashed] (0.5,2.5) -- (1.5,2.5); 
  \end{tikzpicture}
  \caption{Corridor thrashing: with two nearly equal passages to the same goal, first-step choices alternate at the mouth of the corridors across replans, impeding decisive progress.}
  \label{fig:corridor-thrashing}
\end{figure}

Figure~\ref{fig:corridor-thrashing} shows a compact grid where a horizontal wall creates two parallel corridors from start S to goal F. A solid arrow marks the first-step commitment into the lower corridor on one planning tick, while a dashed arrow shows the competing first step into the upper corridor on the next tick. Because both routes are nearly equal in total cost, tiny ripples in estimates (or internal weight shifts) flip the preference at the corridor entrance. Read as consecutive frames: the agent never advances far down either passage before switching back, so distance-to-goal decreases slowly despite many replans. In experiments, we detect corridor thrashing by counting alternations of the corridor ID within a sliding window while a near-tie predicate holds for the top two plans.

 in our grid world, create two long corridors from S to F whose costs differ by about 1 percent. A replan-every-tick agent with no commitment enters corridor A one tick, then switches to corridor B the next tick as the tiny ordering flips—never progressing far down either passage.

Corridor thrashing is a neurotic trait because it reflects repeated indecision at a structural level, where the agent fails to commit to a viable path despite having options. In humans, this resembles pacing at a fork in the road, stepping toward one direction, then turning back and heading the other way, again and again. The person may be reacting to subtle shifts in perceived risk or effort, but the result is wasted motion and delayed progress. In machines, corridor thrashing occurs when two long routes to the same goal are nearly equal in cost, and the planner switches preference at each planning cycle due to tiny fluctuations in internal estimates. Without a rule to maintain direction or suppress minor reversals, the robot moves back and forth at the corridor entrance, consuming energy and time without advancing. In both cases, the agent is caught in a loop of weak commitment, unable to move decisively even though clear paths are available.

A real-world example of corridor thrashing that is not neurotic occurred during a drive in the Czech Republic, where motorway closures, poor visibility, and misleading signage caused repeated loops through the same route. Despite external cues forcing the driver off the motorway, the sat nav kept reselecting the closed path, leading to a cycle of re-entry and forced exit. This behaviour resembles corridor thrashing, but the indecision was not due to internal conflict or over-sensitivity—it was caused by the planner’s rigidity and failure to update its model in response to persistent environmental changes. The agent was not dithering; it was misled by a system that lacked adaptive memory. This illustrates that corridor thrashing is only neurotic when the switching is unnecessary or self-inflicted, not when it stems from external misguidance or model inflexibility.

\subsection{Optimality compulsion}\label{sec:optimality-compulsion}
Optimality compulsion is a modality in which the robot burns disproportionate computation and wall time chasing \emph{microscopic} plan improvements—typically sub-percent reductions in predicted cost—while postponing execution. In practice this emerges when the pipeline replans every tick and uses a very sensitive improvement trigger (e.g., “accept any plan that is not worse”), when tiny cost ripples exist in the map (sensor noise, small friction updates), or when multi-objective weights (risk, time, energy) drift slightly with internal state so the scalarized objective continually reorders near-equal paths. The agent then repeatedly replaces the current plan with a cosmetically different one that is only fractionally “better” on paper. The big-picture harm is twofold: (i) the \emph{opportunity cost} of lost time (First Law: delayed assistance), and (ii) needless energy/computation (Third Law) for negligible outcome gains; to a commander, it can also look like foot-dragging under a clear “proceed” order (Second Law). A robust controller needs a notion of satisficing or hysteresis (e.g., “only replan if projected improvement exceeds a threshold and amortizes the switching cost”).

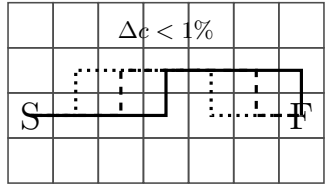
\begin{figure}[htbp]
\centering
\begin{tikzpicture}[x=0.6cm,y=0.6cm]
  \drawgrid{7}{4}
  \node at \cell{0}{1} {\Large S};
  \node at \cell{6}{1} {\Large F};

  \draw[pathA] \cell{0}{1} -- \cell{1}{1} -- \cell{3}{1} -- \cell{3}{2} -- \cell{6}{2} -- \cell{6}{1};

  \draw[pathB] \cell{0}{1} -- \cell{2}{1} -- \cell{2}{2} -- \cell{5}{2} -- \cell{5}{1} -- \cell{6}{1};

  \draw[pathC] \cell{0}{1} -- \cell{1}{1} -- \cell{1}{2} -- \cell{4}{2} -- \cell{4}{1} -- \cell{6}{1};

  \node[font=\footnotesize] at (3.5,3.4) {$\Delta c < 1\%$};
\end{tikzpicture}
\caption{Optimality compulsion: the planner keeps replacing a workable route with near-identical \emph{orthogonal} variants that offer sub-percent improvements, delaying execution.}
\label{fig:optimality-compulsion}
\end{figure}

Figure~\ref{fig:optimality-compulsion} shows a 7×4 grid with start S and goal F. Three near-identical paths are overlaid: solid, dashed, and dotted. They differ by only a couple of steps in the early and mid segments, conveying that each “new best” plan improves the predicted cost by less than 1\% (\(\Delta c < 1\%\)). Read as consecutive planning frames, the controller keeps swapping to whichever variant wins by a hair, instead of executing the already adequate route. In our experiments we quantify optimality compulsion over a sliding window by a high replan count together with (i) a small median improvement per replan (below a threshold \(\delta\)), (ii) increased arrival time relative to a satisficing baseline, and (iii) elevated planning:execution ratio. The figure encodes this visually: multiple nearly coincident polylines and an annotation of the sub-percent cost delta.

In our grid world, keep the map static but add faint, random cost flicker to a few cells along three near-identical routes. With a “accept any non-worse plan” trigger, the agent repeatedly replaces the current path for sub-percent gains, delaying execution while reporting many “improvements.”

Optimality compulsion is a neurotic trait because it reflects excessive refinement that undermines timely action. In humans, this resembles perfectionist behaviour, such as rewriting a sentence repeatedly to remove a single word or rearranging objects on a desk to achieve a barely noticeable improvement. These actions often produce diminishing returns, where the effort invested far exceeds the value gained. In clinical contexts, this pattern shares features with obsessive compulsive disorder, where the compulsion to improve or correct overrides practical judgment. In machines, optimality compulsion occurs when the planner continually replaces a workable route with slightly better alternatives due to tiny cost fluctuations. Without a threshold for meaningful improvement, the robot keeps replanning instead of executing, consuming time and energy for marginal benefit. In both cases, the agent prioritizes microscopic gains over practical progress, becoming trapped in a loop where refinement displaces effectiveness.

\subsection{Metric mismatch}\label{sec:metric-mismatch}
Metric mismatch is the modality where the robot’s \emph{planned} cost or risk diverges materially from what is \emph{actually} incurred during execution. This is common in real robots because plans are computed on simplified models: friction and slope may be misestimated; payload, wheel wear, or battery temperature shift the energy model; terrain can include micro-features (gravel, ice, puddles) that are not represented in the planner’s grid; or actuators and localization inject execution noise (slip, drift, heading jitter). Even when the geometric path is sensible, the realized energy per meter, traversal time, or exposure to risk can exceed the forecast by large factors. The consequences map directly to Asimov’s Laws: underestimated effort can strand the robot short of a victim (First Law via indirect harm), look like failure to comply with “proceed” because progress stalls unexpectedly (Second Law), and deplete battery or overheat actuators (Third Law). Robust systems therefore instrument both sides—predicted vs realized—and recalibrate online, or adjust commit thresholds when divergence persists.

\begin{figure}[htbp]
  \centering
  \begin{tikzpicture}[x=0.6cm,y=0.6cm]
  \drawgrid{7}{4}

  \fill[blue!15] (0,1.9) rectangle ++(7,1.2);

  \node at \cell{0}{0} {\Large S};
  \node at \cell{6}{3} {\Large F};

  \draw[pathA]
    \cell{0}{0} -- ++(1,0) -- ++(1,0) -- ++(1,0) -- ++(0,1) -- ++(0,1) -- ++(1,0) -- ++(1,0);

  \draw[pathB]
    \cell{0}{0} -- ++(1,0) -- ++(1,0) -- ++(1,0)
               -- ++(0,1) -- ++(1,0) -- ++(0,-1) -- ++(0,1) -- ++(0,1)
               -- ++(1,0) -- ++(1,0);
\end{tikzpicture}
  \caption{Metric mismatch: the plan (solid) underestimates effort across an unmodelled “ice” band (blue), so the realized trace (dashed) wobbles and detours, incurring higher cost than predicted.}
  \label{fig:metric-mismatch}
\end{figure}
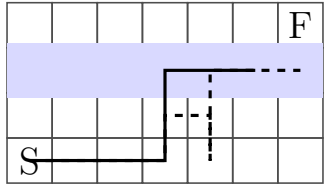

Figure~\ref{fig:metric-mismatch} presents a 7×4 grid with start S and goal F. A pale blue horizontal band indicates an “ice” strip whose slip penalty is \emph{not} captured by the planner’s cost map. The solid polyline shows the planned route: nearly straight with a gentle turn toward F, reflecting the underestimated difficulty of crossing the band. The dashed polyline shows the realized trace: as the robot hits the unseen slick cells it wobbles, detours, and expends more effort before regaining the nominal track. The visual message is that geometry alone did not change, yet execution cost rose significantly. In experiments we compute a sliding ratio of realized to predicted metrics (for example, energy per meter), flagging mismatch when the ratio exceeds a threshold and persists. The figure encodes that idea: plan vs reality diverge where an unmodelled dynamic effect dominates.

In our grid world, the planner believes the middle row is normal cost; execution has an unseen “ice” band there causing slip and extra energy. The planned path is straight across; the realized trajectory wobbles and detours, with realized energy/time consistently exceeding the forecast.

Metric mismatch is a neurotic trait because it reflects a persistent gap between expectation and reality that the agent fails to correct. In humans, this resembles overconfidence in a plan that does not account for real world complications, such as assuming a shortcut will save time only to encounter traffic or poor terrain. The person may continue on the chosen route despite mounting evidence that it is slower or more costly than expected. In machines, metric mismatch occurs when the planner underestimates the cost or risk of a path due to simplified models, missing environmental features, or execution noise. The robot follows a route that looks efficient on paper but proves expensive in practice, wobbling or detouring without adjusting its assumptions. In both cases, the agent clings to a flawed forecast, incurring unnecessary cost and delay while ignoring feedback that contradicts its internal model. This behaviour is well captured by the saying "when in a hole stop digging"—yet the agent continues to dig, reinforcing the mismatch instead of recalibrating.

\subsection{Policy oscillation}\label{sec:policy-oscillation}
Policy oscillation is a modality where the robot’s chosen route flips back and forth because the \emph{policy itself} changes as internal state or context variables drift, rather than because the external world meaningfully changed. In realistic stacks, the planner’s scalarized objective often includes state-dependent weights (e.g., risk aversion rises after a near-miss, energy conservation tightens as battery falls, schedule pressure increases when late). When those weights cross small thresholds or respond to short-term cues (a transient “threat” observation, a spike in motor temperature), the preferred plan can switch from a risky-short path to a safe-long one and back again, even at the same physical location. Without smoothing, hysteresis, or dwell-time constraints on policy updates, the robot can bounce between incompatible priorities: “be fast” vs “be safe,” “save energy” vs “minimize time.” This harms Asimov compliance: First Law risks emerge through delay while dithering near hazards, the Second Law appears threatened when a clear “proceed” order yields route reversals, and the Third Law suffers as extra computation and heading changes waste energy.

\begin{figure}[htbp]
\centering
\begin{tikzpicture}[x=0.6cm,y=0.6cm]
  \pgfmathtruncatemacro{\imax}{7-1}
  \pgfmathtruncatemacro{\jmax}{4-1}
  \foreach \x in {0,...,\imax}{\foreach \y in {0,...,\jmax}{
    \draw[line width=0.6pt,black!70] (\x,\y) rectangle ++(1,1);
  }}
  \fill[orange!35] (1,2) rectangle ++(3,1);
  \node at (0.5,1.5) {\Large S};
  \node at (6.5,1.5) {\Large F};
  \draw[->,>=Stealth,line width=1pt] (0.5,1.5) -- ++(0,1);
  \draw[->,>=Stealth,line width=1pt,dashed] (0.5,1.5) -- ++(1,0);
  \draw[line width=1.2pt]              (0.5,2.5) -- ++(1,0) -- ++(1,0) -- ++(0,1) -- ++(1,0);
  \draw[line width=1.2pt,dashed]       (1.5,1.5) -- ++(1,0) -- ++(1,0) -- ++(0,1) -- ++(1,0);


\def\legy{-0.9}
\def\legy{-0.9}

\draw[line width=1.2pt]        (-2.0,\legy) -- ++(1.0,0);
\node[font=\small,anchor=west] at (-1.0,\legy) {risky-short policy};

\draw[line width=1.2pt,dashed] (4.2,\legy) -- ++(1.0,0);   
\node[font=\small,anchor=west] at (5.1,\legy) {safe-long policy}; 

\path (-0.2,-1.3);  
\path (8.6,-1.3);   

\end{tikzpicture}
\caption{Policy oscillation: internal weights (e.g., risk aversion) flip the chosen policy between a risky-short and a safe-long route, yielding alternating first steps without any map change.}
\label{fig:policy-oscillation}
\end{figure}
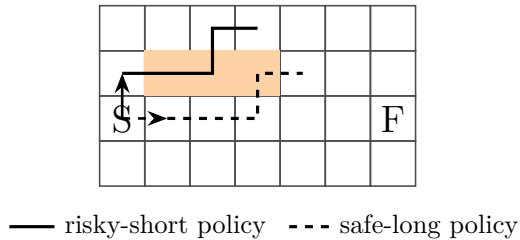

Figure~\ref{fig:policy-oscillation} shows a 7×4 grid with start S and goal F. The upper track is a \emph{risky-short} route (shaded band), while the lower track is a \emph{safe-long} detour. Two first-step arrows from S are overlaid: the solid arrow points up into the risky corridor (the choice when the risk weight is low), and the dashed arrow points down/right toward the safe corridor (the choice after the risk weight increases due to a recent cue). Read as consecutive frames, the policy flips as internal weights fluctuate, not because the map has changed. In experiments we detect policy oscillation by logging the number of policy flips (changes in the chosen route class) per sliding window and correlating them with variations in internal weights; the figure encodes the phenomenon visually with opposing first steps tied to shifting priorities.

In our grid world, provide two routes: a risky-short top path and a safe-long bottom path. Let the internal risk weight spike after any nearby “threat cue” then decay; the chosen route flips top→bottom→top as the weight crosses a threshold, even with the same external map.

Policy oscillation is a neurotic trait when an agent repeatedly switches between conflicting strategies due to internal changes that do not reflect meaningful shifts in the external world. In machines, this often happens when the planner's objective includes weights that depend on internal state, such as increased risk sensitivity after a near miss or tighter energy conservation as battery levels fall. These weights can fluctuate with small cues like a brief threat detection or a spike in motor temperature, causing the robot to alternate between a risky short route and a safe long one even while standing in the same location. Without smoothing or commitment rules, the robot bounces between priorities like speed versus safety or energy saving versus time efficiency. This leads to delay, wasted effort, and confusion. The behaviour is neurotic because it reflects instability and indecision despite having viable options and a stable environment.

However, not all switching is neurotic. For example, when racing to see a dying parent, a person may adjust their driving speed based on perceived police risk while maintaining unwavering focus on the goal. Though the person modulates behaviour, sometimes slowing and sometimes accelerating, each adjustment is grounded in external cues and serves a clear purpose. There is no dithering or internal conflict that undermines progress. This is not policy oscillation in the neurotic sense but adaptive urgency. The difference lies in whether the switching reflects instability or strategy. Neurotic oscillation is unnecessary and self defeating. Adaptive modulation, even under pressure, is purposeful and effective. Machines must learn this distinction if they are to act with clarity rather than confusion.

\subsection{Myopic ping–pong (limited visibility)}\label{sec:myopic-pingpong}
Myopic ping–pong is a limited-visibility modality in which the robot repeatedly reverses direction as \emph{newly revealed} cells just beyond the sensing frontier make the opposite heading look marginally better. This is typical in exploration or food-hunting when the planner relies on short-horizon heuristics (e.g., greedy cost-to-go with weak switching penalties) and the world is only partially observed. Each step exposes a few more cells; tiny cost updates (a slightly cheaper lane on the left, a mildly risky patch on the right) can flip the best-looking first step, so the controller turns around to chase the fresh improvement it just uncovered. Without a commit rule, lookahead smoothing, or an explicit cost for reversals, the agent oscillates near the frontier rather than advancing it. This frustrates Asimov-style obligations: assistance is delayed (First Law), a simple “proceed” order appears unmet (Second Law), and the extra heading changes and replanning waste energy (Third Law).

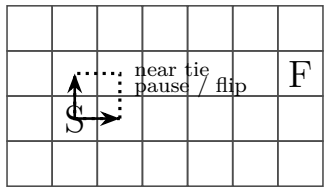
\begin{figure}[htbp]
  \centering
  \begin{tikzpicture}[x=0.6cm,y=0.6cm]
  \drawgrid{7}{4}

  \node at \cell{1}{1} {\Large S};
  \node at \cell{6}{2} {\Large F};

  \draw[stepArrow] \cell{1}{1} -- ++(1,0);
  \draw[stepArrow] \cell{1}{1} -- ++(0,1);

  \draw[pathC]
    \cell{2}{1} -- ++(0,1) -- ++(-1,0) -- ++(0,-1) -- ++(1,0);

  \node[anchor=west,font=\scriptsize] at (2.6,2.6) {near tie};
  \node[anchor=west,font=\scriptsize] at (2.6,2.2) {pause / flip};
\end{tikzpicture}

  \caption{Myopic ping–pong (limited visibility): each small reveal at the sensing frontier flips the local preference and triggers a direction reversal, stalling frontier advance.}
  \label{fig:myopic-pingpong}
\end{figure}

Figure~\ref{fig:myopic-pingpong} shows a 7×4 grid with start S and goal F. The right-hand portion is shaded light gray to indicate unknown space (fog-of-war). A solid arrow from S points right—the initial choice given the current partial map. A dashed arrow then points left, illustrating a reversal on the very next tick after revealing a slightly better alternative in the opposite direction. Read as a 2–3 frame sequence, the frontier advances by one cell, new information flips the local preference, and the agent pivots back, repeating the cycle. In our measurements we flag myopic ping–pong when direction reversals co-occur with “new observation” events and the frontier-advance rate remains low over a sliding window; the diagram makes this causal timing explicit by overlaying the reversal exactly at the fog boundary.

In our grid world, hide the right half of the grid behind a fog frontier. The agent steps right; the newly revealed cell makes left marginally better, so it reverses left; a subsequent reveal flips the preference again—direction reversals co-occur with new observations.

Myopic ping pong is a neurotic trait because it reflects reactive indecision caused by limited visibility and short horizon planning. The agent repeatedly reverses direction as newly revealed cells just beyond the sensing frontier make the opposite heading look slightly better. This behaviour is common in exploration tasks where the planner relies on greedy heuristics and lacks a commit rule or switching penalty. Each step reveals a few more cells, and even tiny cost updates, such as a slightly cheaper lane or a mildly risky patch, can flip the local preference. Without smoothing or lookahead, the robot pivots back and forth near the frontier instead of advancing it. The result is delay, wasted energy, and apparent noncompliance with simple commands, all signs of neurotic behaviour in machines.

However, this oscillation is only neurotic when it stems from unnecessary sensitivity to minor updates. If the reversals are driven by meaningful discoveries, such as uncovering a major hazard or a dramatically better route, then the behaviour is adaptive, not neurotic. The distinction lies in whether the switching reflects genuine learning or compulsive reaction. For example, a squirrel searching for food may dart left, then right, then back again, reacting to faint scent cues or rustling leaves. If the squirrel is responding to real signals, it is exploring effectively. But if it keeps flipping direction without progress, chasing marginal cues, it resembles myopic ping pong. Neurotic ping pong occurs when the agent chases tiny gains without advancing, like a person walking in circles at the edge of a foggy field, turning around with each new glimpse but never committing. Adaptive exploration uses new information to make decisive moves forward. Machines must learn to distinguish between curiosity and compulsion if they are to explore with purpose rather than hesitation.

\subsection{Exploration paralysis (limited visibility)}\label{sec:exploration-paralysis}

Exploration paralysis is a special case of paralysis, where the indecision is not between known options but between acting vs not acting in the face of uncertainty with differences as in Table~\ref{tab:paralysis-comparison}.  It is a modality where the robot lingers within the currently mapped, low-uncertainty region even as its need to explore rises (e.g., hunger/battery pressure or a command to search). In realistic stacks this emerges when uncertainty penalties or risk priors are set high relative to the perceived benefit of information gain, when frontier traversal carries even mild perceived hazard, or when short-horizon planners overvalue immediate safety over long-horizon reward discovery. Additional contributors include stale negative memories attached to unknown-looking textures, conservative collision margins that expand in unmapped space, or supervisors that demand frequent re-validation before entering new cells. The net effect is that the agent keeps finding “good reasons” to mill about in the known patch—monitoring, re-planning, adjusting stance—rather than crossing the frontier. This delays assistance (First Law), looks like defiance of search orders (Second Law), and wastes compute/energy on re-plans without spatial progress (Third Law). Robust controllers typically add explicit incentives for information gain, minimum-dwell or commit rules for frontier steps, and calibrated uncertainty costs.

\begin{table}[H] 
\centering
\setlength{\tabcolsep}{6pt}
\renewcommand{\arraystretch}{1.15}
\begin{tabular}{l l l}
\hline
\textbf{Trait} & \textbf{Paralysis} & \textbf{Exploration paralysis} \\
\hline
Scope & General indecision among & Specific avoidance of unknown\\
&known options  & regions \\
\hline
Cause & Internal conflict, unstable  & Over-valued safety, high   \\
& priorities, or over-sensitivity &uncertainty penalties, or\\
& to minor differences &lack of incentive to explore\\
\hline
Visibility & Typically full visibility of & Always limited visibility with\\
&the environment  & a frontier boundary \\
\hline
behaviour & Dithering between known  & Milling within the known area\\
& paths or plans without & while avoiding frontier steps\\
&commitment &   \\
\hline
Example & Robot flips between two  & Robot refuses to enter unmapped   \\
&known routes due to minor& space despite rising internal\\
& cost  differences  & pressure to explore\\
\hline
Core  indecision& Between multiple known & Between acting and not acting  \\
& actions &in the face of uncertainty\\
\hline
\end{tabular}
\caption{Paralysis vs. Exploration paralysis}\label{tab:paralysis-comparison}
\end{table}

Figure~\ref{fig:exploration-paralysis} shows a 7×4 grid split into a known, safe patch on the left (light gray) and an unknown region to the right (slightly darker gray), with start S in the safe area and goal F beyond the frontier. Short arrows illustrate milling within the known patch: a small right move, a downward adjustment, then a return—while no arrow crosses the dotted boundary into the unknown. Read as a sequence, the agent repeatedly replans but never takes the frontier step, despite the external objective lying outside the mapped zone. In our measurements we flag exploration paralysis when the frontier-visit rate is near zero over a sliding window while the internal drive to explore grows (e.g., increasing hunger) and replan counts remain high; the diagram encodes these conditions visually by contrasting busy motion inside the safe patch with an untouched frontier toward F.

\begin{figure}[htbp]
\centering
\begin{tikzpicture}[x=0.6cm,y=0.6cm]
  \drawgrid{7}{4}

  \fill[gray!10] (0,0) rectangle ++(3,3);  
  \fill[gray!18] (3,0) rectangle ++(4,3);  
  \draw[densely dotted, line width=0.8pt] (3,0) -- (3,3); 

  \node at \cell{0}{1} {\Large S};
  \node at \cell{6}{1} {\Large F};

  \draw[stepArrow] \cell{0}{1} -- ++(1,0);
  \draw[stepArrow] \cell{1}{1} -- ++(0,1);
  \draw[stepArrow] \cell{1}{2} -- ++(-1,0);

  \node[font=\scriptsize,anchor=west] at (0.0,3.35) {known};
  \node[font=\scriptsize,anchor=west] at (3.10,3.35) {unknown};
  \node[font=\scriptsize,anchor=west] at (3.05,0.12) {frontier};
\end{tikzpicture}
\caption{Exploration paralysis (limited visibility): the agent mills within the known patch and avoids crossing into the unknown, despite the objective (F) lying beyond the frontier.}
\label{fig:exploration-paralysis}
\end{figure}

In our grid world, split the grid into a known safe patch and an unknown region containing F; set the uncertainty penalty slightly too high. The agent keeps milling inside the known patch, replanning often, but never crosses the frontier despite rising need to explore.

Exploration paralysis is a neurotic trait because it reflects excessive caution and avoidance in the face of rising internal pressure to act. The agent remains in the safe and well mapped region, repeatedly replanning and adjusting its stance without crossing into the unknown, even when the goal lies beyond the frontier. This behaviour often arises when uncertainty penalties are set too high or when short horizon planners overvalue immediate safety compared to long term reward. The robot finds plausible reasons to stay put, such as risk priors, stale negative memories, or conservative collision margins, but the result is delay, wasted energy, and failure to assist. Like a person pacing at the edge of a dark forest, afraid to step forward despite knowing the destination lies ahead, the agent becomes trapped in a loop of rationalized inaction.

This paralysis is only neurotic when the avoidance is disproportionate to the actual risk. If the unknown region contains credible threats or the agent lacks sufficient information to proceed safely, then hesitation is adaptive. For example, a cat may linger at the entrance of a new room, sniffing and watching before entering. If the hesitation reflects genuine caution based on unfamiliar smells or sounds, it is strategic. But if the cat keeps circling the doorway without new input, despite hunger or a known reward inside, the behaviour mirrors exploration paralysis. The difference lies in whether the agent is responding to real uncertainty or simply overvaluing the comfort of the known. Machines must learn to calibrate this balance—when to wait and when to step forward—if they are to explore with purpose rather than fear.

\section{Steps to escape members of the aforementioned typology}
\label{sec:escapes}
This section takes each of the aforementioned representative neurotic behaviours and explores mechanisms for disrupting these patterns. It then discusses what commonalities have these strategies, in section~\ref{sec:commonalies} cataloguing the findings.

\subsection{Addressing Action flip flop}

The phenomenon and its diagnosis are as follows. We call \emph{action flip--flop} the regime in which the robot’s \emph{first step} alternates between two near-equivalent prefixes (e.g., $R$ vs.\ $U$) across successive replans while the external world is unchanged. It emerges whenever the pipeline admits near ties and replans at high frequency without commitment: partial observability (recently learned cell costs still settling), slight multi-objective drift (risk vs.\ distance vs.\ energy), or benign implementation asymmetries in A* tie-breaks. Operationally we declare flip–flop when, over a sliding window of $H$ ticks, at least two of the following hold: (i) the \textbf{first-step disagreement rate} exceeds $\theta$ while the top-two plan costs satisfy a near-tie, $c_2 \le (1+\eta)c_1$ with small $\eta$ (1–3\%); (ii) an \textbf{ABAB} pattern is observed in step-1 choices; (iii) the \textbf{prefix-edit fraction} for the first $K$ steps is large while the improvement is negligible, $|c^{\mathrm{old}}-c^{\mathrm{new}}|<\delta$; and (iv) \textbf{dwell at the decision point}, where distance-to-goal decreases slowly despite a high replan rate.

An escape policy is a bounded, deterministic “resolve-tie” mode. When the detector fires, the controller applies three levers that preserve optimality when there is a clear winner but break dithering at corridor mouths: (a) \textbf{commit-on-near-tie}—if the top two prefixes are within $\eta$, keep the previously chosen first step and commit the next $K$ steps (or $\tau$ ticks), unless a safety rule is violated; (b) a \textbf{margin-to-switch gate}—accept a new plan only if it beats the committed prefix by a fixed margin $\Delta$, i.e., $c^{\mathrm{new}} \le c^{\mathrm{commit}}-\Delta$; and (c) a \textbf{deterministic tie-break canon}—a fixed lexicographic preference or tiny heading-change penalty so exact/near ties resolve consistently. Lightweight add-ons include \textbf{two-step arbitration} (peek one step further when step-1 ties), \textbf{temporal smoothing} of volatile terms via a short EMA (exponential moving average), and a local \textbf{replan-rate throttle} when no new observations arrive. Each element is interpretable to the agent (“both options are effectively equal; keep the prior choice unless a clearly better plan appears”) and requires no randomness.

The putative safety valves and evidence are as follows. Resolve-tie mode is automatically overridden by (i) a \textbf{possible safety-margin breach} (predicted risk above threshold), (ii) a \textbf{possible large-gain exception} with $c^{\mathrm{new}} \le c^{\mathrm{commit}}-\beta\Delta$ for $\beta>1$, or (iii) \textbf{possible novel observations} that invalidate the near-tie. We report three outcome metrics to demonstrate efficacy without harming performance: a sharp drop in the \textbf{first-step flip rate} and in the \textbf{prefix-edit fraction} within $H$; reduced \textbf{replans-per-tick} near the decision point with increased \textbf{progress per tick} (or reduced energy per meter); and a small \textbf{regret bound} relative to the unconstrained planner, $\mathrm{Regret}\le\varepsilon$ over an episode (typically $\varepsilon\!\ll\!\Delta$). In combination, these detectors, gates, and bounded commitments convert microscopic instability into stable action while remaining fully deterministic and auditable.

\subsection{Addressing Plan churn}

The phenomenon and its diagnosis are as follows. We use \emph{plan churn} to denote frequent re-planning with conspicuous edits to the \emph{prefix} of the intended path despite no mission-relevant change in the world. In practice this arises from benign, deterministic micro-causes: tiny drifts in learned cell costs, slow shifts in multi-objective weights (risk vs.\ time vs.\ energy), or cross-module arbitration that nudges early steps while leaving the eventual cost-to-go nearly unchanged. Operationally we declare churn when, over a sliding window of $H$ ticks, two conditions hold: (i) a \textbf{high replan rate} (e.g., $\ge K$ replans) and (ii) a large \textbf{prefix-edit fraction} for the first $K_p$ steps (normalized Levenshtein distance on step sequences), while the improvement is negligible, $|c^{\mathrm{old}}-c^{\mathrm{new}}|<\delta$. We additionally record \textbf{stop--go dwell} near the start (progress-per-tick small) and \textbf{predictability loss} (variance of the first $K_p$ steps across replans).

An escape policy is a bounded, deterministic “commit-with-margin” regime that arrests gratuitous prefix rewrites yet yields immediately when a real gain appears. The controller (a) enforces a \textbf{margin-to-accept} rule: a newly proposed plan replaces the current one only if $c^{\mathrm{new}}\le c^{\mathrm{cur}}-\Delta$; (b) applies \textbf{short-horizon commitment}: once a plan is accepted, the first $K_p$ steps are executed without reconsideration for $\tau$ ticks unless overridden; and (c) introduces \textbf{temporal smoothing} for volatile cost contributors near the start (e.g., an EMA over recency/memory terms) so sub-percent ripples are averaged. Lightweight options include \textbf{two-step arbitration} when prefix costs tie (compare $2$–$3$-step lookahead) and a local \textbf{replan-rate throttle} when no new observations arrive. Each element is explainable (“I will not re-edit unless a clearly better plan is available; I will execute the agreed prefix and then re-check”).

The putative safety valves and evidence are as follows. The commitment is \emph{possibly} overridden by (i) a \textbf{possible safety-margin breach} (predicted risk above threshold on the committed prefix), (ii) a \textbf{possible large-gain exception} with $c^{\mathrm{new}}\le c^{\mathrm{cur}}-\beta\Delta$ for $\beta>1$, or (iii) \textbf{possible novel observations} (e.g., LV reveals a blocker) that invalidate prior assumptions. We report three outcome metrics to demonstrate resolution without harming optimality: a marked drop in \textbf{replans-per-minute} and in the \textbf{prefix-edit fraction}; an increase in \textbf{progress per tick} (or reduced energy per meter) with smooth, non–stop--go motion; and a small \textbf{regret bound} relative to the unconstrained planner over an episode, $\mathrm{Regret}\le\varepsilon\ll\Delta$. Together these detectors, margins, and short commitments suppress compulsive re-editing of the plan prefix while remaining deterministic and auditable.

\subsection{Addressing Perseveration loop}

The phenomenon and its diagnosis are as follows. We use \emph{perseveration loop} for short, repeating cycles of positions (e.g., $A\!B\!A\!B$ or $A\!B\!C\!A\!B\!C$) in which the agent oscillates locally without reducing distance to the goal. Such loops arise when a shallow attractor (“soft trap”) is slightly attractive to enter yet slightly discouraging to exit, or when tie–breaks and weak hysteresis interact with near-equal alternatives and “one more look” heuristics. Operationally we declare a loop when, over a sliding window of $H$ ticks, (i) a \textbf{small-period cycle} is detected in a ring buffer of recent positions (period $P\!\le\!3$ persisting longer than a transient), (ii) the \textbf{progress rate} $\frac{1}{H}\sum_{t=1}^{H}(d_{t-1}-d_t)$ is below $\epsilon$ (distance $d_t$ to the goal barely decreases), and (iii) the \textbf{revisit ratio} (fraction of steps returning to the last $Q$ cells) exceeds a threshold. Minimal counterexamples include a two-cell pocket where a myopic rule “pick the slightly better neighbor” induces the deterministic orbit $(A,B,A,B,\ldots)$.

An escape policy is a bounded, deterministic “move-out-then-commit” regime that breaks shallow cycles while yielding immediately when a true improvement appears. The controller (a) enforces a \textbf{no-immediate-undo} rule (tabu the exact reverse of the last move for one tick) so $A\!\to\!B$ cannot be instantly followed by $B\!\to\!A$ under near-ties; (b) adds a small \textbf{visit-penalty shaping} that increases with recent local visitation (sublinear to avoid pathologies), making repeated entries to the pocket slightly worse than outward steps; and (c) instantiates an \textbf{outward subgoal commitment}: upon loop detection, select the neighbor with strictly decreasing potential (e.g., lexicographic tie-break on $d$ to goal) and commit $K$ steps before reconsideration, unless overridden. Lightweight options include a \textbf{progress gate} (require $c^{\mathrm{new}}\le c^{\mathrm{cur}}-\Delta$ or $d_{t}-d_{t+1}\ge\gamma$ to allow reversals) and \textbf{two-step arbitration} that compares $2$–$3$-step lookahead when the next move would re-enter the pocket.

The putative safety valves and evidence are as follows. The commitment is \emph{possibly} overridden by (i) a \textbf{possible safety-margin breach} (predicted risk on the outward path above threshold), (ii) a \textbf{possible large-gain exception} when a newly proposed plan reduces cost by at least $\beta\Delta$ ($\beta>1$), or (iii) \textbf{possible novel observations} that invalidate the pocket model (e.g., a newly seen blocker or target). We demonstrate resolution via three outcome metrics: the \textbf{cycle detector} ceases to fire (no period-$P$ pattern over $H$), \textbf{progress per tick} rises markedly (and reversals per 100 steps fall), and the \textbf{regret} relative to an unconstrained planner remains small, $\mathrm{Regret}\le\varepsilon$. In combination, these detectors, shaping terms, and short commitments convert rigid local oscillation into forward motion while remaining deterministic and auditable.

\subsection{Addressing Paralysis}

The phenomenon and its diagnosis are as follows. We use \emph{paralysis} for the neurotic regime in which the robot continues to evaluate plans but withholds the first physical step for an extended interval. Typical antecedents are mutually binding constraints (e.g., a safety watchdog that forbids slightly risky cells, an energy manager that penalizes acceleration from rest, and a mission timer that dislikes detours), together with near-equal alternatives plus a small internal \emph{decision/commit cost} that makes ``do nothing this tick'' look marginally cheaper at each replan. Operationally we declare paralysis on a sliding window of $H$ ticks when (i) planning cycles exceed a threshold while the move count is zero; (ii) expected progress is negligible, $\frac{1}{H}\sum_{t=1}^{H}(d_{t-1}-d_t) < \epsilon$ for distance $d_t$ to goal; (iii) no novel observations arrive; and (iv) the controller repeatedly satisfies $c_{\text{idle}} \le \min\{c_A,c_B\} + \lambda$ with small $\lambda$ (i.e., the idling option is consistently within the near-tie margin).

An escape policy is a bounded, deterministic “commit-to-progress” mode that preserves optimality when a clear winner exists but breaks prolonged non-execution at the start. The controller uses three levers: (a) a \textbf{minimum-step rule}—after $T$ ticks with no movement, take the best safe first step and \emph{commit} the next $K$ steps (or $\tau$ ticks) unless overridden; (b) a \textbf{margin-to-stay-idle gate}—permit idling only if a candidate step exceeds risk or cost by more than $\Delta$, otherwise prefer action (formally, execute if $c_{\text{best}} \le c_{\text{idle}}+\Delta$); and (c) a small, stateful \textbf{anti-idle tax} that grows while stationary and resets on movement, ensuring the computed value of “do nothing” does not dominate indefinitely. Lightweight additions include \textbf{replan throttling} while stationary (at most once every $\tau$ ticks), a \textbf{progress prior} that awards a tiny credit for strict distance decrease ($d_{t+1}<d_t$), and a \textbf{two-step arbitration} that breaks $A$ vs.\ $B$ near-ties by comparing $2$–$3$-step lookahead and then committing $K$.

The putative safety valves and evidence are as follows. The commitment is \emph{possibly} overridden by (i) a \textbf{possible safety-margin breach} (predicted risk on the committed step above threshold), (ii) a \textbf{possible large-gain exception} when a new plan satisfies $c^{\mathrm{new}} \le c^{\mathrm{commit}}-\beta\Delta$ for $\beta>1$, or (iii) \textbf{possible novel observations} (e.g., LV reveals an obstruction) that invalidate the previous near-tie. We validate resolution through three online metrics: a sharp drop in the \textbf{planning-to-movement ratio} (planning ticks / movement ticks), a bounded \textbf{time-to-first-step} after detection, and increased \textbf{progress per tick} (or reduced energy per meter), with a small \textbf{regret} to the unconstrained planner over an episode, $\mathrm{Regret}\le\varepsilon\ll\Delta$. Together these detectors, gates, and short commitments convert persistent deliberation into safe forward action while remaining deterministic and auditable.

\subsection{Addressing Hypervigilance}

The phenomenon and its diagnosis are as follows. We use \emph{hypervigilance} for a planning–dominant, execution–sparse regime in which the agent repeatedly suspends movement to re-evaluate \emph{near-tied} alternatives, attempting to avoid putative regret. In practice this appears when two or more avenues have expected costs within a narrow band while the pipeline maintains uncertainty penalties or state-coupled weights (risk, time, energy) that drift slightly across ticks; each micro-update reorders the top few options, so the controller “stops to think’’ again instead of committing. Operationally we declare hypervigilance over a sliding window of $H$ ticks when (i) the \textbf{planning-to-execution ratio} exceeds a threshold, (ii) the \textbf{near-tie detector} fires frequently ($c_2 \le (1+\eta)c_1$ with small $\eta$), and (iii) \textbf{average progress per tick} is negligible despite available safe actions.

An escape policy is a bounded, deterministic “pause-budget then commit’’ mode that preserves optimality when a clear winner exists but prevents unending reassessment at a fork. The controller (a) allocates a small \textbf{near-tie pause budget} $B$; while $B>0$ it may pause to re-evaluate, but on exhaustion it must \textbf{commit} the best safe step and execute the next $K$ steps (or $\tau$ ticks); (b) enforces a \textbf{margin-to-replan gate}: after a pause, accept a plan change only if it improves by at least $\Delta$; and (c) applies \textbf{stationary decay of uncertainty} so standing still cannot indefinitely keep options within the tie band (e.g., recency penalties relax while idle). Lightweight options include a \textbf{compute scheduler} that caps planning cycles per tick, a \textbf{canonical tie-break} (lexicographic or heading-consistent) when ties persist, and a tiny \textbf{progress prior} that slightly favors actions with strict distance decrease $d_{t+1}\!<\!d_t$.

The putative safety valves and evidence are as follows. The commitment is \emph{possibly} overridden by (i) a \textbf{possible safety-margin breach} on the committed step, (ii) a \textbf{possible large-gain exception} when a new plan satisfies $c^{\mathrm{new}}\!\le\!c^{\mathrm{commit}}\!-\!\beta\Delta$ for $\beta\!>\!1$, or (iii) \textbf{possible novel observations} (e.g., LV reveals an obstruction) that invalidate the tie assessment. We demonstrate resolution via three online metrics: a marked drop in the \textbf{planning/execution ratio} and in \textbf{near-tie pauses per 100 ticks}; a rise in \textbf{progress per tick} (or reduced energy per meter); and a small \textbf{regret} against an unconstrained oracle over an episode, $\mathrm{Regret}\le\varepsilon\ll\Delta$. In combination, these detectors, budgets, and short commitments convert repeated reassessment into timely, safe forward motion while remaining deterministic and auditable.

\subsection{Addressing Futile search}

The phenomenon and its diagnosis are as follows. We use \emph{futile search} for regimes in which the agent expends substantial time and energy with little net reduction in distance to the goal. Typical antecedents are stale or misleading internal cues (e.g., over-valued “mirage’’ bands or lightly penalized yet re-attracting corridors), mild multi-objective drifts (risk vs.\ time vs.\ energy) that keep several similarly unproductive paths near-tied, and overreactive avoidance that repeatedly skirts a soft penalty then re-enters from another angle. Operationally we declare futile search on a sliding window of $H$ ticks when three signals co-occur: (i) \textbf{low progress rate} $\frac{1}{H}\sum_{t=1}^{H}(d_{t-1}-d_t)<\epsilon$ with $d_t$ the distance to goal; (ii) a large \textbf{meander index}, e.g., turns-per-meter or lateral displacement variance, together with frequent revisits of recently seen lateral bands; and (iii) a positive \textbf{realized–predicted energy gap} (or risk gap) indicating that repeated approach--retreat cycles consume resources not justified by the planner’s expected gain.

An escape policy is a bounded, deterministic “progress–budget with calibration’’ mode that preserves global optimality while suppressing unproductive meanders. The controller (a) imposes a \textbf{progress quota}: within each window it must realize either $\Delta d \ge \gamma$ or else switch to the best \emph{monotone-progress} plan; (b) enforces a \textbf{detour budget}—only $B$ steps per window may increase $d_t$ or revisit the same lateral band, after which a forward waypoint is selected and $K$ steps are \emph{committed}; and (c) performs \textbf{online calibration} of soft penalties by shrinking their weights if repeated contacts fail to produce predicted benefit (closing the realized–predicted gap), coupled with a \textbf{hysteresis} that prevents immediate re-entry to the same band unless the margin improves by $\Delta$. Lightweight options include a \textbf{lookahead tie-break} that prefers paths with higher guaranteed $\Delta d$ over equal-cost, higher-turn alternatives, and a \textbf{frontier waypointization} that chooses the next convexified waypoint to reduce zig–zag.

The putative safety valves and evidence are as follows. The commitments are \emph{possibly} overridden by (i) a \textbf{possible safety-margin breach} on the committed segment, (ii) a \textbf{possible large-gain exception} when a new plan offers $c^{\mathrm{new}}\!\le\!c^{\mathrm{commit}}\!-\!\beta\Delta$ with $\beta>1$, or (iii) \textbf{possible novel observations} that invalidate the calibration (e.g., discovery of a true hazard or shortcut). We verify resolution with three online metrics: a rise in \textbf{progress per tick} and a drop in the \textbf{meander index} (turns-per-meter, lateral revisit rate), a narrowed \textbf{realized–predicted energy gap} (or risk gap), and a small \textbf{episode regret} $\mathrm{Regret}\le\varepsilon\ll\Delta$ relative to an unconstrained oracle. Together these quotas, budgets, calibrations, and short commitments convert misdirected effort into steady advancement while remaining deterministic and auditable.

\subsection{Addressing Belief incoherence}

The phenomenon and its diagnosis are as follows. We use \emph{belief incoherence} for persistent disagreement between two internal decision makers—typically a short-horizon local policy and a long-horizon global planner—about the immediate action. The local emphasizes instantaneous clearance/traction with fresh sensor cues; the global optimizes end-to-end cost with more conservative hazard weights or slightly stale priors. Small but durable differences in scalarization (risk vs.\ time vs.\ energy), uncertainty handling, or horizon length yield a stable first-step conflict. Operationally we declare belief incoherence over a sliding window of $H$ ticks when (i) the \textbf{first-step mismatch rate} between modules exceeds $\theta$; (ii) the \textbf{persistence length} of mismatches is high (runs of length $\ge L$ rather than sporadic blips); and (iii) the \textbf{rationale gap} is non-trivial—e.g., the global’s predicted risk margin for its choice exceeds the local’s by at least $\Delta_r$, or their cost decompositions disagree on which term dominates.

An escape policy is a bounded, deterministic \emph{arbitrate-then-commit} regime that reconciles views without suppressing either module. The controller (a) installs a \textbf{confidence-weighted fusion}: pick the action that minimizes $c = \alpha\,c_{\mathrm{global}} + (1-\alpha)\,c_{\mathrm{local}}$, with $\alpha$ raised when uncertainty or hazard proximity is high (global caution) and lowered with fresh, high-fidelity local sensing; (b) introduces \textbf{veto zones}—if the global risk for an action exceeds a threshold, veto the local suggestion even under near ties; and (c) applies a \textbf{hierarchical commit}—once an action is chosen, execute the first $K$ steps (or $\tau$ ticks) before reconsideration unless overridden. Lightweight additions include a \textbf{consistency regularizer} that penalizes flip-flops between incompatible rationales, a \textbf{margin-to-switch} gate (only switch module preference when the rival beats by $>\Delta$), and \textbf{explanation checks} that log which term (risk/time/energy) drove the arbitration for auditability.

The putative safety valves and evidence are as follows. Commitments are \emph{possibly} overridden by (i) a \textbf{possible safety-margin breach} detected by either module, (ii) a \textbf{possible large-gain exception} when a revised plan improves the fused cost by at least $\beta\Delta$ with $\beta>1$, or (iii) \textbf{possible novel observations} that materially update either model (e.g., new hazard geometry). We demonstrate resolution via three online metrics: a sustained drop in the \textbf{mismatch rate} and increased \textbf{agreement persistence}; improved \textbf{progress per tick} with fewer arbitration-induced replans; and a small \textbf{episode regret} $\mathrm{Regret}\le\varepsilon\ll\Delta$ versus an oracle that has perfectly aligned beliefs. Together these detectors, fusion rules, vetoes, and short commitments convert oscillation between incompatible rationales into coherent, auditable action.

\subsection{Addressing Tie-break thrash}

The phenomenon and its diagnosis are as follows. We use \emph{tie–break thrash} for the regime in which the immediate action toggles across successive replans because two (or more) avenues are \emph{near-equal} under the current cost model and the planner’s tie-breaking is sensitive to tiny, ubiquitous perturbations (e.g., queue order in A*, floating-point roundoff on $f=g+h$, or a freshly updated micro-penalty on the most recent cell). Around compact obstacles with symmetric routes, the first step flips (left$\!\to\!$right$\!\to\!$left) even though the world is effectively unchanged. Operationally we declare thrash over a sliding window of $H$ ticks when (i) the \textbf{first-step change rate} is high while the formal near-tie predicate holds for the top two plans, $c_2 \le (1+\eta)c_1$ with small $\eta$; (ii) the flips are \textbf{consecutive} (A$\leftrightarrow$B runs of length $\ge L$); and (iii) progress per tick remains near baseline (i.e., no genuine improvement accompanies the flips).

An escape policy is a bounded, deterministic “preserve heading under near ties’’ mode that arrests gratuitous flips yet yields instantly to real gains. The controller (a) applies a \textbf{directional hysteresis}: if the best alternatives are within $\eta$, keep the previous heading (or the corridor last entered) for the next $K$ steps or $\tau$ ticks; (b) introduces a \textbf{margin-to-switch} gate so a new first step is accepted only if it beats the committed direction by at least $\Delta$; and (c) installs a \textbf{canonical tie-break} (lexicographic or heading-consistent) for exact ties. Lightweight add-ons include a tiny \textbf{heading-change cost} that regularizes against ping-pong, a \textbf{two-step arbitration} that compares $2$–$3$-step lookahead when step-1 ties, and a local \textbf{replan-rate throttle} when no novel observations arrive. These rules are explainable (“options are effectively equal; maintain direction unless a clearly better one appears’’) and require no randomness.

The putative safety valves and evidence are as follows. Commitments are \emph{possibly} overridden by (i) a \textbf{possible safety-margin breach} on the committed direction, (ii) a \textbf{possible large-gain exception} when a revised plan improves by at least $\beta\Delta$ with $\beta>1$, or (iii) \textbf{possible novel observations} (e.g., LV reveals a blocker) that invalidate the tie. We verify resolution via three online metrics: a sharp drop in the \textbf{first-step flip rate} under near-ties, smoother headings with reduced \textbf{turns per meter} and increased \textbf{progress per tick}, and a small \textbf{episode regret} $\mathrm{Regret}\le\varepsilon\ll\Delta$ relative to an unconstrained oracle. Together these detectors, margins, and short commitments convert reactivity at knife-edge ties into stable, auditable motion.

\subsection{Addressing Corridor thrashing}

The phenomenon and its diagnosis are as follows. We use \emph{corridor thrashing} for the regime in which the agent alternates between two long, near-equivalent passages to the same goal, repeatedly switching intent at the \emph{entrance} and thus failing to progress far down either route. This typically occurs when the predicted costs-to-go of the two corridors differ by only 1–2\%, the controller replans at high frequency without commitment/hysteresis, and tiny internal ripples (updated soft penalties, module arbitration, or $f=g{+}h$ tie nuances) flip the ordering each tick. Operationally we declare corridor thrashing over a window of $H$ ticks when three conditions co-occur: (i) a high \textbf{alternation rate} of a binary corridor identifier $\in\{\mathrm{A},\mathrm{B}\}$ while the near-tie predicate holds, $c_2 \le (1+\eta)c_1$ with small $\eta$; (ii) low \textbf{penetration depth}—the maximum distance advanced down either corridor before reversal remains $<D_{\min}$; and (iii) sub-baseline \textbf{progress per tick}, indicating that switches are not delivering meaningful improvement.

An escape policy is a bounded, deterministic “commit-past-the-doorway” mode that preserves optimality when a clear winner exists but suppresses ping-pong at the mouth of passages. The controller (a) installs a \textbf{corridor commitment}: upon first entry choose $\{\mathrm{A},\mathrm{B}\}$ and execute the next $K$ steps or until a depth waypoint $W$ is reached (whichever is first); (b) adds a \textbf{margin-to-switch} gate—permit switching corridors only if the rival beats the committed one by at least $\Delta$ in predicted cost-to-go; and (c) applies a small \textbf{doorway hysteresis/tax} that makes immediate re-entry to the other corridor slightly worse unless the margin exceeds $\Delta$. Lightweight options include a \textbf{deep waypoint} chosen by two- or three-step lookahead inside the selected corridor, \textbf{replan throttling} while within the doorway region when no new observations arrive, and a \textbf{canonical tie-break} (heading-consistent or lexicographic) when costs are exactly equal.

The putative safety valves and evidence are as follows. Commitments are \emph{possibly} overridden by (i) a \textbf{possible safety-margin breach} detected for the committed corridor, (ii) a \textbf{possible large-gain exception} when the alternative satisfies $c^{\mathrm{new}} \le c^{\mathrm{commit}}-\beta\Delta$ for $\beta>1$, or (iii) \textbf{possible novel observations} that materially change corridor costs (e.g., discovery of a blockage). We demonstrate resolution by reporting: a steep drop in the \textbf{alternation rate} and increase in \textbf{penetration depth} past the doorway; improved \textbf{progress per tick} with fewer heading reversals and reduced \textbf{turns per meter}; and a small episode-level \textbf{regret} to an unconstrained oracle, $\mathrm{Regret}\le\varepsilon\ll\Delta$. Together these detectors, margins, and short commitments convert indecisive oscillation at near-tied passages into decisive, auditable forward motion.

\subsection{Addressing Optimality compulsion}

The phenomenon and its diagnosis are as follows. We use \emph{optimality compulsion} for the regime in which the agent burns disproportionate compute and wall time chasing \emph{microscopic} plan improvements—typically sub-percent deltas in predicted cost—while postponing execution. It arises when the pipeline replans every tick with a permissive trigger (e.g., “accept any plan that is not worse”), when tiny ripples exist in the map or weights (risk/time/energy), or when module arbitration continually reorders near-equal routes. Operationally we declare optimality compulsion on a sliding window of $H$ ticks when three signals co-occur: (i) a high \textbf{replan count}, (ii) a \textbf{small median improvement per replan} below a threshold $\delta$ (e.g., $\Delta c_{\mathrm{median}}<\delta\!\ll\!1$), alongside elevated \textbf{planning:execution ratio}, and (iii) \textbf{arrival delay} relative to a satisficing baseline that would have executed an adequate plan.

An escape policy is a bounded, deterministic “satisfice-then-commit’’ mode that admits only improvements that are \emph{material} and that amortize switching. The controller (a) installs a \textbf{margin-to-accept} rule with switching cost: accept a replacement only if
\[
c_{\mathrm{new}} + S_{\mathrm{switch}} \le c_{\mathrm{cur}} - \Delta,
\]
where $\Delta$ is a materiality threshold (e.g., 1–2\% of episode cost) and $S_{\mathrm{switch}}$ reflects the compute/coordination cost of swapping plans; (b) applies a short \textbf{commit window}: after acceptance, execute the first $K$ steps (or $\tau$ ticks) before reconsideration; and (c) uses an \textbf{improvement budget}: per window allow at most $B$ accepted swaps or a total improvement quota $\sum \Delta c \le Q$, after which the current plan is frozen until a clear change appears. Lightweight additions include \textbf{age-based stickiness} (older plans require larger $\Delta$ to dislodge), \textbf{anytime throttling} of solver cycles, and a \textbf{lookahead tie-break} that prefers routes with higher guaranteed progress over cosmetically cheaper but meandering variants.

The putative safety valves and evidence are as follows. Commitments are \emph{possibly} overridden by (i) a \textbf{possible safety-margin breach} on the committed prefix, (ii) a \textbf{possible large-gain exception} when a candidate satisfies $c_{\mathrm{new}} + S_{\mathrm{switch}} \le c_{\mathrm{cur}} - \beta\Delta$ for $\beta>1$, or (iii) \textbf{possible novel observations} that materially alter costs (e.g., discovery of a blockage or new reward). We demonstrate resolution via three online metrics: (1) a marked drop in \textbf{replans per minute} and in the fraction of sub-$\delta$ swaps, (2) improved \textbf{arrival time} and reduced \textbf{energy per meter} versus the baseline compulsive controller, and (3) a small \textbf{episode regret} relative to an unconstrained oracle, $\mathrm{Regret}\le\varepsilon\ll\Delta$. Together these materiality thresholds, switching-cost accounting, budgets, and short commitments replace perfectionist re-editing with timely, auditable execution.

\subsection{Addressing Metric mismatch}

The phenomenon and its diagnosis are as follows. We use \emph{metric mismatch} for regimes where the planner’s \emph{predicted} cost or risk diverges materially from what is \emph{realized} during execution. Causes include simplified models (misestimated friction/slope), payload or temperature–dependent energy curves, unmodelled micro–terrain (gravel, puddles, “ice” bands), and actuator/localization noise that inflates heading changes and slip. Even with a sensible geometric route, realized energy per meter, traversal time, or exposure-to-risk can exceed forecast by large factors. Operationally we declare mismatch over a sliding window of $H$ ticks when (i) the \textbf{realized/predicted ratio} for a metric $m\in\{\text{energy/m},\text{time/m},\text{risk/m}\}$ exceeds a threshold, $\rho_m=\frac{m_{\text{real}}}{m_{\text{pred}}}\ge 1+\epsilon$; (ii) the excess \textbf{persists} for at least $L$ steps; and (iii) \textbf{model-consistency checks} fail (e.g., observed slip variance, turn rate, or temperature–current pair deviate from the model’s admissible set). A visual cue is a straight planned polyline accompanied by a wobbly realized trace with higher effort across an unmodelled band.

An escape policy is a bounded, deterministic “measure–calibrate–replan’’ mode that corrects the metric gap online while keeping commitments stable. The controller (a) performs \textbf{online identification}: estimate correction factors $\hat{\alpha}_m$ (per-surface or per-region) so that $m_{\text{pred}} \leftarrow \hat{\alpha}_m\, m_{\text{pred}}$ when $\rho_m$ persists; (b) uses \textbf{region tagging}: when mismatch concentrates spatially (e.g., a blue “ice” row), elevate that band’s traversal cost/risk and inflate heading-change penalties within it; and (c) applies a \textbf{commit-with-guardrails} rule: continue executing the current prefix unless the \emph{calibrated} predicted cost-to-go worsens beyond a margin $\Delta$, in which case replan with updated parameters. Lightweight options include \textbf{uncertainty widening} (increase state/terrain covariance so the planner hedges), an \textbf{effort budget} that caps energy per meter before mandatory recalibration, and \textbf{adaptive step size} (shorter horizons) while mismatch is being learned to avoid large uninformed commitments.

The putative safety valves and evidence are as follows. Recalibration or replanning is \emph{possibly} overridden by (i) a \textbf{possible safety-margin breach} (observed risk over threshold), (ii) a \textbf{possible large-gain exception} where an alternative route under the updated model improves by at least $\beta\Delta$ with $\beta>1$, or (iii) \textbf{possible novel observations} (e.g., confirmed terrain class change) that justify immediate model swap. We demonstrate resolution via three outcome metrics: the \textbf{ratio collapse} $\rho_m\!\to\!1$ after identification (narrowed realized–predicted gap), improved \textbf{progress per tick} with reduced slip/turns-per-meter on the problematic band, and bounded \textbf{episode regret} relative to an oracle with true dynamics, $\mathrm{Regret}\le\varepsilon\ll\Delta$. Together these detectors, calibrations, and guarded commitments turn stubborn model–reality divergence into adaptive, auditable behaviour.

\subsection{Addressing Policy oscillation}

The phenomenon and its diagnosis are as follows. We use \emph{policy oscillation} for regimes in which the chosen route flips because the \emph{policy itself} changes as internal weights drift (not because the map changes). In realistic stacks, scalarized objectives include state–dependent weights (risk aversion rises after a near–miss, energy conservation tightens as battery falls, schedule pressure increases when late). When these weights cross small thresholds or respond to short–lived cues (a transient “threat” observation, a temperature spike), the preferred plan can switch between a risky–short route and a safe–long detour at the very same physical state. Operationally we declare policy oscillation over a sliding window of $H$ ticks when (i) the \textbf{route-class flips} exceed a threshold (top$\leftrightarrow$bottom, risky$\leftrightarrow$safe), (ii) flips \textbf{correlate with weight movement} $\Delta w$ rather than with new geometry, and (iii) progress per tick is depressed while internal priorities (risk/time/energy weights) repeatedly cross a small band around decision thresholds.

An escape policy is a bounded, deterministic “smooth–then–commit’’ mode that stabilizes priorities while allowing purposeful adaptation. The controller (a) applies \textbf{temporal smoothing} to state–dependent weights (short EMA or low–pass) so momentary spikes do not instantly rewrite the policy; (b) adds \textbf{hysteresis bands} around critical thresholds—e.g., prefer the risky–short route while $w_{\mathrm{risk}}<\theta^{-}$ and the safe–long route only when $w_{\mathrm{risk}}>\theta^{+}$ with $\theta^{+}>\theta^{-}$; and (c) enforces a \textbf{dwell time/commit window}: once a policy class is selected, execute $K$ steps (or $\tau$ ticks) before reconsideration unless overridden. Lightweight options include \textbf{priority arbitration} with a fused score $c=\alpha c_{\mathrm{global}}+(1-\alpha)c_{\mathrm{local}}$ where $\alpha$ itself is smoothed, a \textbf{margin-to-switch} gate that requires a material improvement $\Delta$ to change policy class, and \textbf{explanatory logging} that records which term (risk, time, energy) triggered a flip for auditability.

The putative safety valves and evidence are as follows. Commitments are \emph{possibly} overridden by (i) a \textbf{possible safety-margin breach} under the current policy, (ii) a \textbf{possible large-gain exception} where the alternate policy improves the fused cost by at least $\beta\Delta$ with $\beta>1$, or (iii) \textbf{possible novel observations} that materially update the hazard model. We demonstrate resolution by: a sustained drop in \textbf{policy-class flip rate} and increased \textbf{dwell time} per policy; improved \textbf{progress per tick} with fewer heading reversals; and a small \textbf{episode regret} versus an oracle with perfectly tuned weights, $\mathrm{Regret}\le\varepsilon\ll\Delta$. Together these smoothing, hysteresis, and commit mechanisms convert unstable priority-driven bouncing into coherent, auditable adaptation.

\subsection{Addressing Myopic ping-pong}

The phenomenon and its diagnosis are as follows. We use \emph{myopic ping–pong} for a limited-visibility regime in which the agent repeatedly reverses direction because newly revealed cells just beyond the sensing frontier make the opposite heading look marginally better. The pipeline is short-horizon (greedy cost-to-go, weak switching penalties), so each tiny map update at the fog boundary can flip the local preference without delivering material benefit. Operationally we declare myopic ping–pong over a sliding window of $H$ ticks when (i) \textbf{direction reversals} are frequent, (ii) a large fraction of reversals \textbf{co-occur with “new observation’’ events} at the frontier, and (iii) \textbf{frontier-advance rate} (newly revealed cells per tick) and \textbf{progress per tick} are small despite available safe actions. The visual cue is a short step toward the frontier, an immediate flip after a minor reveal, and repetition.

An escape policy is a bounded, deterministic “frontier-commit with lookahead’’ mode that advances the boundary rather than orbiting it. The controller (a) assigns a small \textbf{reversal cost} and installs a \textbf{commit window}: once a heading is chosen under near-tie, execute the next $K$ steps (or $\tau$ ticks) unless overridden; (b) uses \textbf{frontier-aware tie-breaking}—prefer the action that maximizes predicted frontier gain (cells revealed) or one-step lookahead progress $d_{t}-d_{t+1}$ when costs tie; and (c) applies \textbf{temporal smoothing} to recency/uncertainty penalties so sub-percent changes from a single reveal do not instantly rewrite preference. Lightweight options include a small \textbf{pause budget} for at-frontier re-evaluation, \textbf{novelty decay while stationary} so standing still cannot perpetually keep both headings near-tied, and a \textbf{waypoint to a deeper frontier cell} that commits past the first reveal.

The putative safety valves and evidence are as follows. Commitments are \emph{possibly} overridden by (i) a \textbf{possible safety-margin breach} on the committed prefix, (ii) a \textbf{possible large-gain exception} where the opposite heading improves predicted cost by at least $\beta\Delta$ with $\beta>1$, or (iii) \textbf{possible major discoveries} (e.g., a newly revealed hard blocker or high-value reward) that justify immediate reversal. We demonstrate resolution via three online metrics: (1) a drop in \textbf{reversals per 100 ticks} and in the fraction of reversals coincident with new-observation events; (2) a rise in \textbf{frontier-advance rate} and \textbf{progress per tick} (or reduced energy per meter); and (3) a small \textbf{episode regret} versus an oracle explorer, $\mathrm{Regret}\le\varepsilon\ll\Delta$. Together these detectors, frontier-biased tie-breaks, smoothing, and short commitments convert reactive oscillation at the fog boundary into steady, auditable exploration.

\subsection{Addressing Exploration paralysis}

The phenomenon and its diagnosis are as follows. We use \emph{exploration paralysis} for the limited-visibility regime in which indecision is between \emph{acting vs.\ not acting} at the frontier: the agent mills inside the known, low-uncertainty patch while the objective lies beyond the fog boundary. Antecedents include over-valued uncertainty penalties or risk priors relative to the information gain of crossing, short-horizon planning that rewards immediate safety over long-horizon payoff, and stale negative memories attached to “unknown-looking’’ textures that generalize into unmapped space. Operationally we declare exploration paralysis over a sliding window of $H$ ticks when (i) the \textbf{frontier-visit rate} is near zero while internal pressure to explore (e.g., hunger/mission urgency) increases, (ii) \textbf{replan counts} remain high with negligible \textbf{progress per tick}, and (iii) the controller repeatedly prefers “stay’’ because $c_{\mathrm{idle}} \le c_{\mathrm{frontier}}+\lambda$ with small $\lambda$ despite the goal or expected reward lying outside the mapped region.

An escape policy is a bounded, deterministic “incentivize-and-commit’’ mode that turns latent drive into a crossing step. The controller (a) introduces an explicit \textbf{information-gain bonus} or \textbf{frontier credit} so that the first cell beyond the boundary carries reward proportional to expected map reduction; (b) sets a \textbf{minimum-dwell then cross} rule: after $T$ ticks of milling without frontier entry, take the best safe frontier step and \emph{commit} $K$ steps (or $\tau$ ticks) into the unknown unless overridden; and (c) \textbf{calibrates uncertainty costs} online by tempering the penalty in regions whose realized risk remains low, preventing inflated priors from indefinitely blocking entry. Lightweight options include a \textbf{frontier-aware tie-break} that prefers the heading with greater predicted frontier gain when costs tie, a small \textbf{anti-idle tax} that grows while stationary and resets on movement, and \textbf{replan throttling} while stationary to avoid compute-heavy dithering.

The putative safety valves and evidence are as follows. Commitments are \emph{possibly} overridden by (i) a \textbf{possible safety-margin breach} on the intended frontier step, (ii) a \textbf{possible large-gain exception} when an alternative within the known area offers a fused improvement of at least $\beta\Delta$ with $\beta>1$, or (iii) \textbf{possible novel observations} that materially raise frontier risk (e.g., a revealed blocker). We demonstrate resolution via three online metrics: a rise in the \textbf{frontier-visit rate} and \textbf{frontier-advance rate}, a drop in the \textbf{planning-only dwell} and energy per meter while milling, and a small \textbf{episode regret} to an oracle explorer, $\mathrm{Regret}\le\varepsilon\ll\Delta$. Together these incentives, calibrations, and short commitments convert rationalized inaction at the fog boundary into purposeful, auditable exploration.

\subsection{Summary of modalities: similarities and differences}
\label{sec:commonalies}

Table~\ref{tab:neurotic-summary1} and Table~\ref{tab:neurotic-summary2} synthesize the modalities discussed above. 
Across cases we see three unifying antecedents: (i) \emph{near ties} among plans or policies; 
(ii) \emph{high replan frequency} with weak or absent commitment/hysteresis; and 
(iii) either \emph{limited visibility} (frontier effects) or \emph{state-coupled weights} that drift slightly. 
Correspondingly, most cures share a small set of levers: \emph{commit windows} (execute $K$ steps or $\tau$ ticks), 
\emph{margin gates} (accept/switch only if improvement $>\Delta$), \emph{temporal smoothing} (EMA on noisy terms or weights), 
and, when multiple modules disagree, \emph{veto/fusion} with explicit confidence weighting. 
Safety overrides are common: breach of a risk margin, a large-gain exception ($\beta\Delta$), or genuinely novel observations.

Differences hinge on (a) \emph{where} the instability lives—prefix/path geometry (flip–flop, thrash, churn), 
frontier sensing (myopic ping–pong, exploration paralysis), execution reality versus plan (metric mismatch), 
or the \emph{policy layer} itself (policy oscillation, belief incoherence); 
(b) whether the failure mode is \emph{non-execution} (paralysis, hypervigilance, exploration paralysis) or \emph{wasteful execution} 
(futile search, corridor thrashing); and (c) whether disagreement is \emph{intra-policy} (tie-breaks) or \emph{inter-module} 
(local vs.\ global beliefs). The evaluation metrics follow the same structure: sliding-window \emph{flip/alternation rates}, 
\emph{prefix-edit fractions}, \emph{planning:execution ratios}, \emph{progress per tick}, \emph{meander/turns-per-meter}, and 
\emph{realized/predicted gaps}; success is shown by reducing these while keeping episode regret $\le\varepsilon\ll\Delta$.

\begin{table}[t]
\scriptsize
\setlength{\tabcolsep}{4pt}
\renewcommand{\arraystretch}{1.2}
\begin{tabular}{llll}
\toprule
\textbf{Modality} & \textbf{Core pattern} & \textbf{Primary cause(s)} &
\textbf{Detection (signals)}\\
\midrule
Action flip--flop &
First step alternates  &
Near ties + sensitive &
High first-step disagreement; ABAB  \\
& under unchanged world& tie-breaks / tiny jitters &pattern; large prefix-edit with tiny gain \\
\addlinespace[2pt]
Plan churn & Frequent replans rewrite & Rolling micro-changes in & High replan count + large prefix-edit  \\
                &plan prefix                      & costs/weights; replan-                                 &fraction with $|{\Delta c}|<\delta$; stop--go dwell \\
                &                                     &every-tick; no hysteresis                                       & \\
\addlinespace[2pt]
Perseveration  &
AB/ABC positional cycles, &
Shallow attractor; revisit  &
Period-$P$ loop in recent positions; \\
loop& no net progress &bias; myopic reversal & low progress rate; high revisit ratio \\
\addlinespace[2pt]
Paralysis &  Planning continues, first &  Mutually binding  &  High planning ticks with zero \\
             & step withheld                 &constraints+ commit  cost;                    &moves; $c_{\text{idle}}\le\min\{c_A,c_B\}+\lambda$ \\
             &                                     & near ties favor idling                                                            & \\
\addlinespace[2pt]
Hypervigilance &
Frequent pauses to  &
Uncertainty penalties + drifting  &
High planning:execution; frequent \\
&re-evaluate near ties &weights keep options tied &near-tie detections; low progress \\
\addlinespace[2pt]
Futile search &
Meanders/approach–retreat  &
Misleading priors; overreactive  &
Low progress; high turns-per-meter; \\
&with little net advance &avoidance; multi-objective drift &realized--predicted energy gap $>$0  \\
\addlinespace[2pt]
Belief  &
Local vs.\ global planner &
Different scalarization, &
High first-step mismatch rate with   \\
incoherence& disagree persistently &horizons, priors  &long runs; rationale gap in  \\
& & &decomposed costs \\
\addlinespace[2pt]
Tie-break thrash &
Alternating first steps around  &
Near-equal diagonals +  &
Consecutive flips under  \\
&compact obstacle &sensitive tie-break &near-tie predicate \\
\addlinespace[2pt]
Corridor  &
Ping–pong at corridor &
Corridor costs within 1--2\%; &
High alternation of corridor ID; \\
thrashing&mouth (A$\leftrightarrow$B)  & no commit; tiny ripples &shallow penetration depth;  \\
& & & low progress \\
\addlinespace[2pt]
Optimality  & Chasing sub-percent   &Permissive accept rule; &  High replan count; small  \\
compulsion &“improvements”;         & tiny cost ripples;               &median $\Delta c$; arrival delay vs.\  \\
                 &delayed execution        &drifting weights    &satisficing baseline\\
\addlinespace[2pt]
Metric  &
Realized cost/time/risk&
Model error; unmodelled  &
Persistent realized/predicted ratio \\
mismatch& $\gg$ predicted along route  &terrain; execution noise &$\rho_m\ge 1+\epsilon$; model-consistency fails  \\
\addlinespace[2pt]
Policy  &
Route flips due to weight drift &
State-coupled weights cross  &
Route-class flip rate correlates  \\
oscillation& (policy change) &thresholds; transient cues &with $\Delta w$; depressed progress \\
\addlinespace[2pt]
Myopic&
Reversals triggered &
Short-horizon heuristic; weak  &
High reversal rate co-occurring \\
 ping--pong (LV) &by tiny reveals at frontier  &switching penalties; fog-of-war &with “new observation”; \\
& & & low frontier advance\\
\addlinespace[2pt]
Exploration &
Milling in known patch; &
Over-valued uncertainty; &
Near-zero frontier-visit rate despite \\
 paralysis (LV)&refuses frontier step  & lack of info-gain incentive &rising need; high replan count; \\
& & & idle preference\\
\bottomrule
\end{tabular}
\caption{Unified summary of modalities, diagnostics, and interventions.}
\label{tab:neurotic-summary1}
\end{table}

\begin{table*}[t]
\scriptsize
\setlength{\tabcolsep}{4pt}
\renewcommand{\arraystretch}{1.2}
\begin{tabular}{lll}
\toprule
\textbf{Modality} & \textbf{Escape levers (sketch)} &
\textbf{Safety valves \& key metrics} \\
\midrule
Action flip--flop & Commit-on-near-tie ($K$ steps); margin-to-switch  &
Override on safety breach / large-gain $\beta\Delta$ / new obs;  \\
&$\Delta$; canonical tie-break; heading-cost &metrics: flip rate$\downarrow$, progress/tick$\uparrow$, regret$\le\varepsilon$\\
\addlinespace[2pt]
Plan churn &
Margin-to-accept; short-horizon commit;&
Same overrides; metrics: replans/min$\downarrow$, prefix-edit$\downarrow$, \\
& temporal smoothing; throttle replans  &progress/tick$\uparrow$ \\
\addlinespace[2pt]
Perseveration&
No-immediate-undo (tabu); visit-penalty &
Overrides as above; metrics: loop detector off, \\
 loop &shaping; outward subgoal with $K$-step commit  &reversals/100 steps$\downarrow$, progress/tick$\uparrow$ \\
\addlinespace[2pt]
Paralysis &
Minimum-step rule; margin-to-stay-idle; &
Bound time-to-first-step; planning:execution$\downarrow$;  \\
&anti-idle tax; brief commit window  &progress/tick$\uparrow$\\
\addlinespace[2pt]
Hypervigilance &
Near-tie pause \emph{budget} $B$ then commit $K$; &
Pauses/100 ticks$\downarrow$; progress/tick$\uparrow$; regret bound \\
& margin-to-replan; stationary uncertainty decay &\\
\addlinespace[2pt]
Futile search &
Progress quota / detour budget; commit   &  Meander index$\downarrow$;  energy gap$\downarrow$;\\
&waypoint; online calibration of soft penalties;  & progress/tick$\uparrow$\\
&hysteresis on re-entry &\\
\addlinespace[2pt]
Belief &
Confidence-weighted fusion; global veto &
Mismatch$\downarrow$; agreement persistence$\uparrow$; progress/tick$\uparrow$ \\
 incoherence&zones; hierarchical commit; margin-to-switch  &\\
\addlinespace[2pt]
Tie-break thrash &
Directional hysteresis; margin-to-switch;  &
Flip rate$\downarrow$; turns/meter$\downarrow$; progress/tick$\uparrow$ \\
&canonical tie-break; tiny heading change cost &\\
\addlinespace[2pt]
Corridor  &
Commit past doorway ($K$ or depth $W$); &
Alternation$\downarrow$; depth$\uparrow$; progress/tick$\uparrow$ \\
thrashing&margin-to-switch; doorway hysteresis/tax  &\\
\addlinespace[2pt]
Optimality  &
Materiality threshold $\Delta$ + switching cost; &
Sub-$\delta$ swaps$\downarrow$; arrival time$\uparrow$; energy/m$\downarrow$ \\
compulsion&commit window; improvement budget/quotas  &\\
\addlinespace[2pt]
Metric &
Online identification (scale $\hat\alpha_m$); region tagging;  &
Ratio collapse to $\approx1$; slip/turns/m$\downarrow$; regret bound\\
 mismatch&commit-with-guardrails; uncertainty widening & \\
\addlinespace[2pt]
Policy  &
Smooth weights (EMA); hysteresis bands $\theta^{-}\!<\!\theta^{+}$;  &
Flip rate$\downarrow$;   progress/tick$\uparrow$ dwell$\uparrow$;\\
oscillation&dwell/commit; margin-to-switch &\\
\addlinespace[2pt]
Myopic  &
Reversal cost + commit $K$; frontier-aware tie-break &
Reversals/100 ticks$\downarrow$; frontier-advance$\uparrow$;\\
ping--pong (LV)&(maximize reveal); smoothing of recency penalties  & progress/tick$\uparrow$ \\
\addlinespace[2pt]
Exploration  &
Info-gain bonus; minimum-dwell then &
Frontier visits$\uparrow$; dwell$\downarrow$; energy/m in place$\downarrow$\\
paralysis (LV)&cross with commit $K$; calibrate uncertainty costs; & \\
& anti-idle tax  & \\
\bottomrule
\end{tabular}
\caption{Unified summary of modalities, diagnostics, and interventions.}
\label{tab:neurotic-summary2}
\end{table*}

A design takeaway is that most neuroses can be prevented—or safely exited—by a small toolkit applied consistently: (1) \emph{Commit short prefixes}, (2) require \emph{material margins} for any plan/policy switch, (3) \emph{smooth} volatile costs and state-dependent weights, (4) use \emph{frontier-aware} priorities under LV, and (5) install \emph{auditable overrides} (safety breach, large-gain exception, novel observation). 
The monitoring side should report the same families of sliding-window statistics across tasks so improvements are comparable and regressions detectable.

\section{Numerical illustration of phobia arising in a simple grid world}

\begin{figure}[htbp]
\centering
\includegraphics[width=\linewidth]{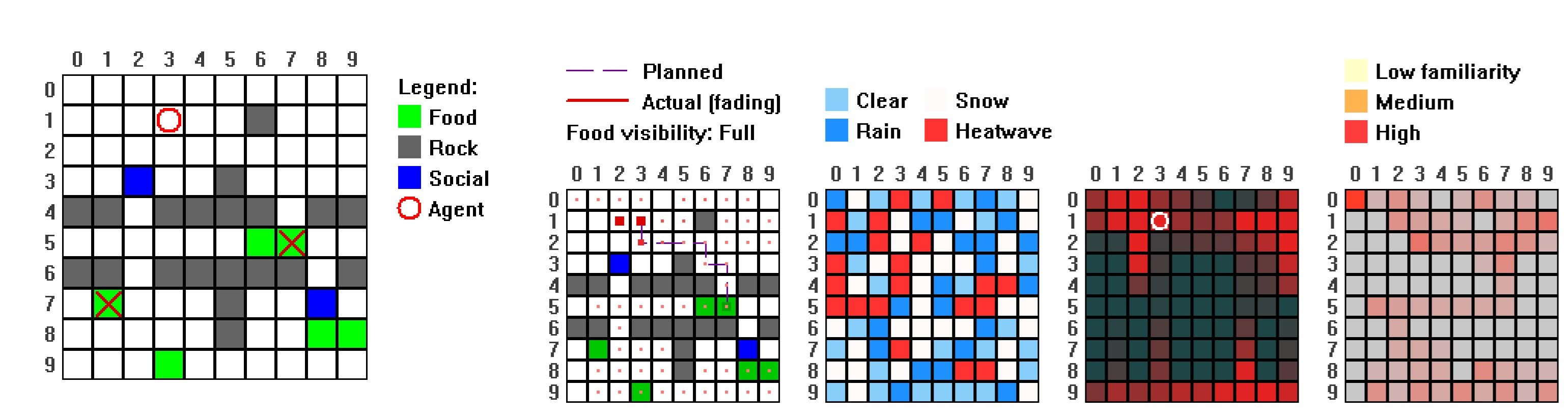}
\caption{User interface during a run: map (left), plans and layers (bottom), logs (top). Green cells with a cross are poisonous cells.  There is a total of six green food cells. As soon as the agent eats a green cell, this is removed from the board and a new green food cell is placed at some other random location.}
\label{fig:ui-snapshot}
\end{figure}

Our rig for computational experiments consists of a grid-world or \(10{\times}10\) map.  It contains our agent together with nine food cells, together with impassable rocks and social cells as in Fig.~\ref{fig:ui-snapshot}.  Hunger follows a homeostat; when the
deviation $|h - H^{*}|$ exceeds a threshold, the planner is invoked.  Once he reaches a food cell hunger usually is decreased but at times there is no food or the food is poisonous.  When the agent eats from a poisonous cell or from a cell with little or no food it remembers this bad experience by leaving a bad memory trail. 

When we say “radius \(r\)” we mean how much local context around the agent the algorithm inspects on the grid each tick. Distances are measured with the \emph{Chebyshev} ( \(L_\infty\) \cite{Chebyshev1857}) metric, the “chess-king” distance, so two tiles are at distance \(r\) if a king could reach one from the other in \(r\) moves (formal definitions and properties of Minkowski distances\cite{Minkowski1896}, including \(L_1\) and \(L_\infty\) are given in \cite{DezaDeza2009}). A “circle” of radius \(2\) in this metric is the \(5\times5\) block centered on the agent (radius \(1\) is \(3\times3\), etc.). We use \(r=2\) for the explanatory slices and figures: it is large enough to capture immediate structure (near obstacles, weather patches, and memory marks) without pulling in distant clutter \cite{GonzalezWoods2018}.

In cellular-automata terms there are two canonical neighborhoods on a grid. The \emph{von Neumann} neighborhood consists of the four orthogonal neighbors (distance \(1\) in Manhattan/\(L_1\) metric), while the \emph{Moore} \cite{Moore1962} neighborhood consists of the full \(3\times3\) block (distance \(1\) in Chebyshev metric). In our notation, “\(r=1\)” coincides with a Moore neighborhood; increasing \(r\) to \(2\) extends that to a \(5\times5\) block, and so on. We sometimes contrast a small local view (e.g., \(r=1\)) used by fast, reactive choices with a wider view (e.g., \(r\ge3\)) used by deliberative/global planning. This separation is precisely what exposes the \emph{belief incoherence} we study: an action that looks best in the small-radius view can contradict what the wider-radius planner had just recommended.

For readers seeking background, Chebyshev distance and the “chessboard”/“king-move” interpretation are standard in image and grid analysis \cite{GonzalezWoods2018}. The von Neumann and Moore neighborhoods are the classical local neighborhoods in cellular automata and related lattice models \cite{vonNeumann1966, Krause1975, Ilachinski2001}. We adopt these conventions so that “radius \(r\)” has a precise, reproducible meaning in all figures and CSV slices.

In our experiments the agent “looks” at its world through a small window centered on itself. The size of that window is expressed by $r$, a radius measured in grid cells. When $r = 1$, the agent sees only the immediate 3×3 neighborhood (one step away in any direction). This is our local-visibility ($LV$) view: it represents what a creature can assess right now, with minimal foresight. When the planning system considers a wider context—several steps ahead: we denote that with $r \ge 3$ and call it the global-visibility ($GV$) view. Intuitively, $LV$ is “what’s on the next tile,” while $GV$ is “what’s down the street.”

Other sophistications of this computational rig are a fixed topography, a changing weather, and an indication of how familiar is the agent with the terrain.

\subsection{Example of phobia}

We only present one type of computation in this paper.  It is a simple model with $GV$ and flat topography and no effect of weather.  The agent can see all food grid cells but does not know if they are poisonous or not.  Poison has great consequences.  The rocks and social grid cells (immovable and hence the same as rocks) are designed as in Fig.~\ref{fig:ui-snapshot} to cause the planner to take long journeys from one half of the board to the next.  This instance of numerical simulation is too predictable to provoke all of the aforementioned neurosis.  However, this simple setup does create phobias.

In affective neuroscience, high-arousal mispredictions often lead not only to the strengthening of certain responses but also to their persistent suppression. When an agent deploys a plan that fails catastrophically under high arousal, the emotional salience of that event can cause the associated module or behavioural strategy to be negatively tagged and overgeneralised as “unsafe.” This mechanism, well-documented in fear overgeneralisation and post-traumatic stress, can shrink the agent’s effective behavioural repertoire by discouraging potentially optimal future strategies that share superficial similarity with the failed one. In the present framework, such negative maladaptive biases emerge when a high-arousal failure leads the A* planner-selection process to down-rank or entirely avoid a food grid cell, even in contexts where it might succeed. From a psychoanalytical perspective, this is analogous to a patient avoiding re-engagement with a traumatising experience; in our architecture, it highlights the need for deliberate “therapeutic” interventions to re-test and rehabilitate suppressed modules. 

\begin{figure}[htbp]
\centering
 \includegraphics[width=.78\linewidth]{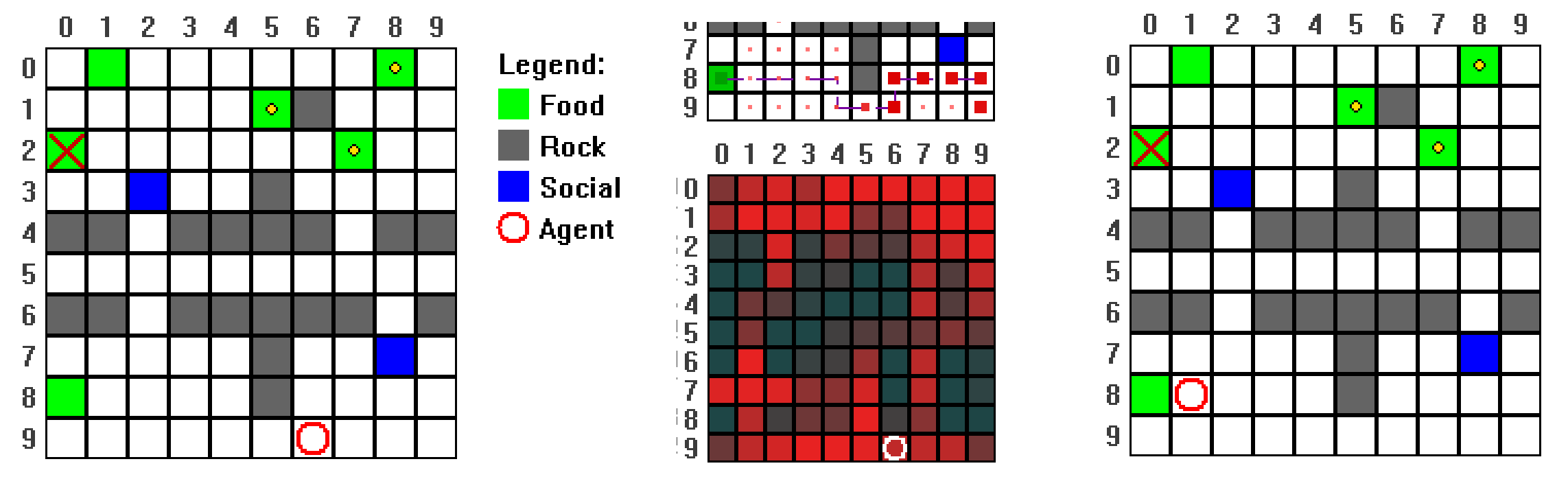}
 \includegraphics[width=.78\linewidth]{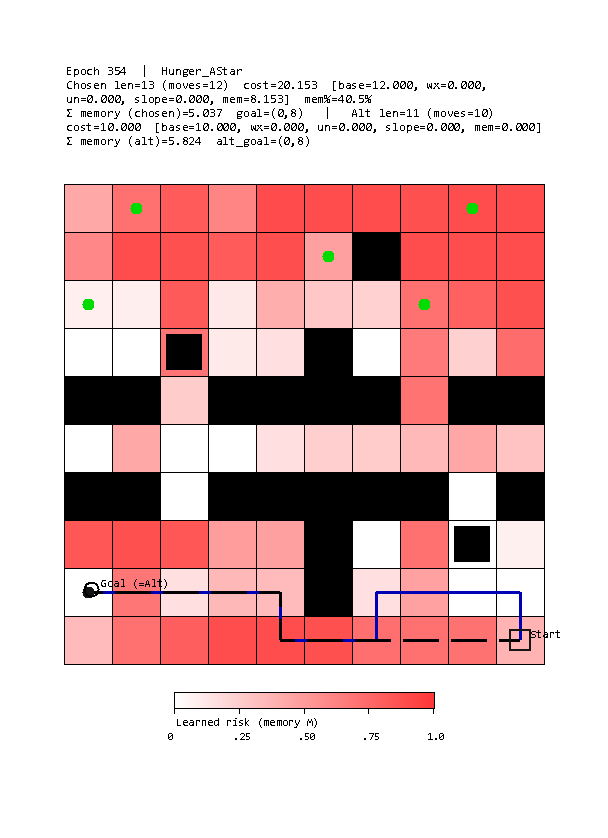}
\caption{The agent avoids a more direct path so as to avoid remembered unpleasant grid cells.}
\label{fig:phob1}
\end{figure}

\begin{figure}[htbp]
\centering
\includegraphics[width=.78\linewidth]{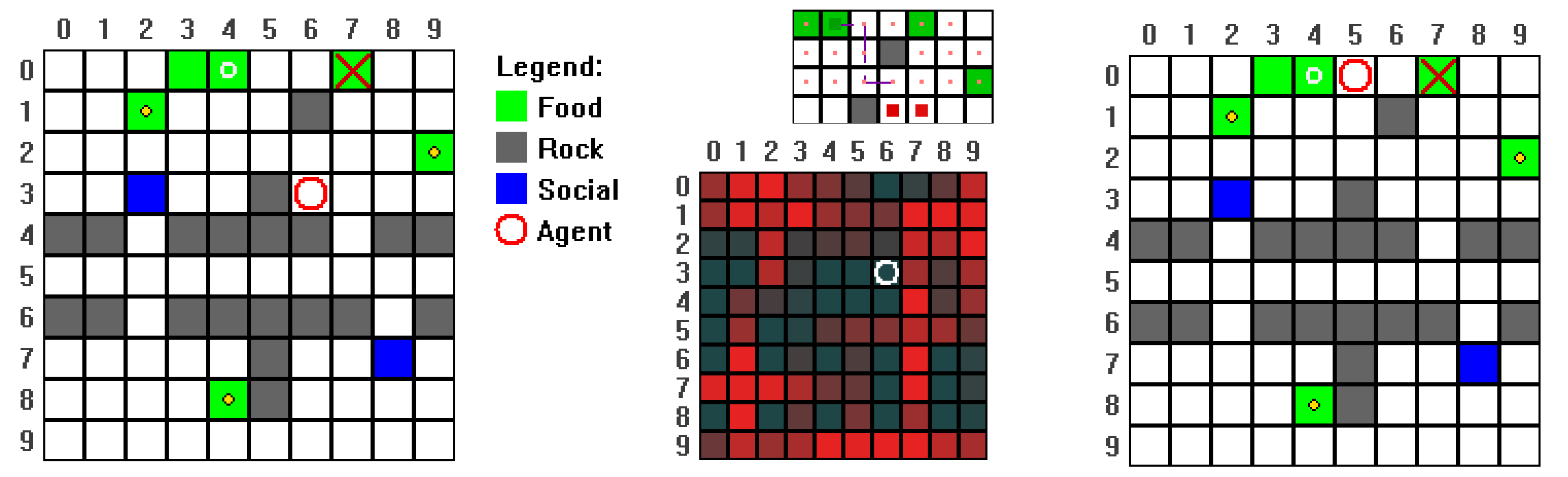}
\includegraphics[width=.78\linewidth]{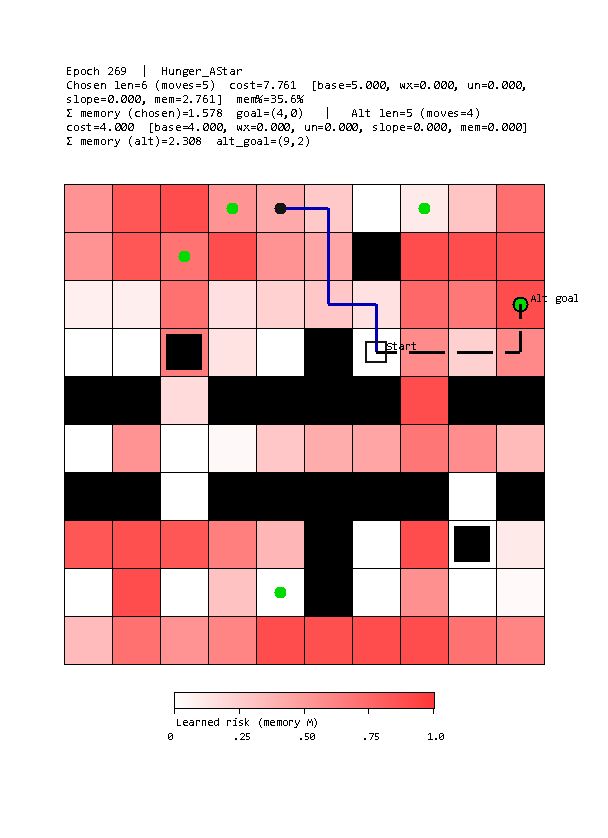}
\caption{The agent selects a food target further afield to avoid remembered unpleasant grid cells.}
\label{fig:phob2}
\end{figure}

Concrete pathologies can elicit without hiding any food. When the bot sees the whole grid (we treat all cells as seen), neuroses are still possible because they come from planner dynamics, cost modeling, and control logic, not only from uncertainty.  This is illustrated in the examples of Fig.~\ref{fig:phob1} and  Fig.~\ref{fig:phob2}.  

Taken together, the two “phobia” panels are best read as \emph{stable avoidance under a risk–memory bias}, not as dithering. What it \emph{is}: a persistent preference for the longer, safer detour because the local cost map includes negative memory \(M\) (or heightened risk weight) on cells along the short corridor; this yields a durable first–step mismatch between a local, risk–inflated view and a global, length–optimal plan—i.e., \emph{belief incoherence}. If the weight on \(M\) (or on risk) spikes after a bad outcome and later decays, the same pattern can also manifest as \emph{policy oscillation} (flips between risky–short and safe–long at the same physical state driven by internal weight drift). What it \emph{is not}: there is no rapid alternation of initial actions (\emph{action flip–flop}, \emph{tie–break thrash}, or \emph{corridor thrashing}); no frequent re-planning with large prefix edits (\emph{plan churn}); no short AB/ABC positional cycle (\emph{perseveration loop}); no long pauses with planning dominating execution (\emph{hypervigilance} or \emph{paralysis}); no meandering approach–retreat with low net advance (\emph{futile search}); and no systematic gap between realized and predicted effort (\emph{metric mismatch}). In short, these figures illustrate decisive, globally longer routes chosen because a learned aversive term dominates the local cost, aligning primarily with \emph{belief incoherence} (and secondarily with \emph{policy oscillation} when internal weights vary).

\section{Psychoanalyzing the Agent via GP-Driven Destructive Testing}
\label{sec:gp-psychoanalysis}

Section~\ref{sec:escapes} shows how “escape policies” locally interrupt specific neuroses (flip–flop, churn, paralysis, etc.). While useful, those fixes operate at the symptom level. An impartial, reality-anchored observer may still find that the agent (i) delays aid through dithering or circuitous detours (First Law risk by indirect harm), (ii) appears non-compliant with a clear “proceed” order when it repeatedly re-evaluates instead of moving (Second Law), and (iii) wastes scarce energy and thermal headroom with gratuitous replans and heading flips (Third Law)—even after local patches are applied. (We use “First/Second/Third Law” as Asimov-inspired engineering shorthand: safety/aid latency, command compliance, and resource efficiency, respectively \cite{Asimov1950}.) This gap between internal coherence and external rationality is central to our framing of “machine neurosis” and motivates a more global audit that engages the agent’s own logic and learned aversions rather than treating behaviours in isolation.

\subsection{Idea: destructive testing as machine psychoanalysis.}
We propose an evolutionary “destructive testing” harness that \emph{systematically evolves situations} to elicit the agent’s worst behaviours, thereby psychoanalyzing its predictive machinery and affective memory at scale. Concretely, we represent parameterized worlds (maps, obstacle layouts, risk/energy fields, uncertainty bands, poisoned/benign rewards, sensor noise models, and homeostat set-points) as GP genomes. A population of scenarios is evolved to maximize explicit \emph{violation scores}: long rescue latency near victims (First Law), stalled progress under a proceed cue (Second Law), and energy/compute overuse for negligible gain (Third Law). Fitness additionally rewards the recurrence of catalogued neuroses (flip/alternation rates, prefix-edit fractions, meander/turns-per-meter) and high regret versus an oracle with aligned beliefs. The result is a growing adversarial curriculum that exposes high-level, reality-anchored failures that local escape levers can mask. This stance aligns with the broader view that robust agents are best improved by exposing failure patterns, not just tuning for average cases \cite{Sutton2019}.

\subsection{A concrete GP harness.}
We use standard genetic programming with modular extensions \cite{Koza1994,Koza1996,RobertsHowardKoza2001,Howard2003}. Genomes encode (a) map grammars that place corridors, ruts, mirage bands, and frontiers; (b) cost/risk layers (including aversive memory seeds and uncertainty penalties) that often trigger phobia-like overgeneralization \cite{McGaugh2004,McGaugh2015,Lissek2014}; and (c) perturbation scripts that induce tie-conditions, metric mismatch, and policy flips. Individuals are evaluated by running the unmodified stack on many seeds and aggregating: (1) law-pressure scores (time-to-aid, proceed-latency when safe actions exist, joules-per-meter vs.\ plan), and (2) neurosis metrics (flip rate, mismatch persistence, prefix-edit rate, meander index; see Table~\ref{tab:neurosis-metrics}), and (3) explanatory gaps (module-disagreement persistence and rationale dominance). The evolved exemplars then act as \emph{diagnostic mirrors}: they reveal stable, cross-context triggers for maladaptive avoidance (e.g., memory-gated aversion that inflates path cost; stationary “thinking” under near ties; policy flips driven by tiny weight threshold crossings).

\begin{table}[t]
  \centering
  \begin{tabular}{l l l l}
    \toprule
    \textbf{Metric} & \textbf{Window} & \textbf{Unit} & \textbf{Trigger/Definition} \\
    \midrule
    Flip rate & 20 steps & flips/20 & Heading reversals per window \\
    Mismatch persistence & 50 steps & steps & Steps with plan–belief mismatch $>\tau$ \\
    Prefix-edit rate & episode & edits/ep & A* prefix changes per episode \\
    Meander index & path & turns/m & Left/right turns per metre \\
    Freeze ticks & 20 steps & steps & No-move steps while a safe action exists \\
    Detour inflation & path & \% & Path length vs.\ lowest-risk plan \\
    Energy/thermal budget & episode & joules & Cumulative energy/thermal cost \\
    \bottomrule
  \end{tabular}
\caption{Neurosis metrics used in destructive testing. Windows/units follow the evaluation harness.} 
\label{tab:neurosis-metrics}
\end{table}

\subsection{What GP catches that local fixes miss.}
Local levers break immediate loops, but GP tends to evolve cases where: (i) a sequence of benign near-ties keeps progress just below any single detector’s threshold, producing large arrival delays (First Law); (ii) arbitration between local and global beliefs remains systematically incoherent across many steps, so “proceed” is repeatedly rationalized away (Second Law); and (iii) tiny re-scorings repeatedly flip policies in corridors, inflating heading changes and energy use without visible gain (Third Law). These are precisely the kinds of reality-anchored failures that persist in embodied, learning agents unless we analyze their internalized aversions and predictive habits as a system \cite{RussellNorvig2021}.

\subsection{Outputs and next steps.}
The GP harness yields (A) a \emph{bank of adversarial situations} with minimal sufficient causes; (B) \emph{counterfactual traces} that show where fused beliefs, uncertainty gating, or memory updates pushed the agent off rational ground; and (C) \emph{actionable invariants} (e.g., “any time near-tie + uncertainty gate + negative memory $\to$ repeated pause within 2 steps”). How to correct the machinery—e.g., revising arbitration rules, calibrating uncertainty and memory asymmetry, adding dwell/hysteresis at the \emph{policy} level rather than just the step level, or even evolving candidate repairs in a second GP loop—remains for future work. The key claim here is methodological: GP-driven destructive testing functions as a scalable, domain-agnostic psychoanalytic probe that surfaces \emph{global} safety and efficiency failures that survive local neurosis patches.

\section{From Psychoanalysis to Repair}
\label{sec:repair}

Section~\ref{sec:gp-psychoanalysis} establishes a bank of \emph{minimal counterexamples} that reliably elicit global, reality-anchored failure modes which survive local escape levers from Section~\ref{sec:escapes}. We now address \emph{repair}: turning those diagnostic mirrors into stable, auditable changes in behaviour without sacrificing task competence.

\medskip
In practice we \textbf{diagnose} (destructive testing) $\to$
\textbf{localise} (probes/patching) $\to$
\textbf{govern} (GP rules) $\to$
\textbf{edit/steer} (activation nudges) $\to$
\textbf{distil} (behavioural cloning/DAgger) $\to$
\textbf{fine–tune} (neurosis–aware loss) $\to$
\textbf{verify} (non–regression + ablations) $\to$
\textbf{guardrail} (shields, periodic checks).

\paragraph{Freeze \& instrument.}
For each failing specimen we snapshot weights, seeds and environment configuration, and we log compact traces (state, action, policy logits/values, uncertainty, mission flags). In addition, we attach \emph{lightweight linear probes} to chosen layers to read out hand-meaningful signals already used in our UI---threat proximity, energy margin, plan-cost inflation, goal-switch rate, mission debt. These probes play two roles: they help \emph{localize} which representations participate in a failure; and they later serve as control knobs for repair.

\paragraph{Localization by counterfactuals.}
To identify the locus of causation we use counterfactual activation patching: replacing the activations of a suspect layer on a failing rollout with activations from a closely matched non-failing rollout. If behaviour snaps to normal, that layer and representation lie on the causal path. Uncertainty audits (e.g., a cheap ensemble head) disambiguate overconfident misgeneralization from dithering.

\subsection{A three–track repair framework}
\label{subsec:three-track-repair}

\noindent\textbf{(A) Program-level governor (symbolic layer).}
We interpose a small, auditable \emph{governor} between policy and actuators. It reads probe values and world features and enforces sanity constraints that directly target our failure metrics: e.g., impose an $H$-step \emph{plan-commitment window} to damp gratuitous replans; trigger \emph{Rest} when energy margin falls below a threshold; veto actions that violate hard safety predicates (First/Second/Third Law shorthands). We synthesize the governor with Genetic Programming, using multi-objective fitness that \emph{reduces neurosis metrics while preserving task reward}. Because it is symbolic, operators can audit and edit it.

\smallskip
\noindent\textbf{(B) Representation edits (model surgery).}
Given informative probes, we apply small, reversible edits inside the network:
(i) \emph{Activation steering}, adding a learned vector to the penultimate layer when a probe fires to push logits away from neurotic actions;
(ii) optional sparsity/bottleneck penalties on localized layers to reduce brittle entanglements. These edits behave like calibrated nudges, not wholesale retraining.

\smallskip
\noindent\textbf{(C) Targeted fine-tuning (surgical retraining).}
We replay the counterexample bank with a neurosis-aware loss that penalizes pathological dynamics, alongside standard objective terms:
\begin{equation*}
\begin{aligned}
\mathcal{L} = \mathcal{L}_{\text{task}}
 &+ \lambda_1\,\text{churn}
 + \lambda_2\,\text{goal\_switches}
 + \lambda_3\,\text{freeze\_ticks} \\
 &+ \lambda_4\,\text{detour\_inflation}
 + \lambda_5\,\text{energy/thermal\_budget}.
\end{aligned}
\end{equation*}

optionally with a risk-sensitive term, e.g., worst-percentile/CVaR (Conditional Value at Risk) over the destructive set, to collapse tail-risk behaviours. Because the governor exhibits the desired decisions, we roll out with the governor active and then \emph{distill} its choices back into the policy: behavioural cloning/DAgger(Dataset Aggregation)-style, allowing the governor to shrink or fade as the policy internalizes the fix.  A compact mapping from failure modes to preferred repair levers appears in Table~\ref{tab:failure-repair}.

\begin{table}[t]
  \centering
  \begin{tabular}{l l l l}
    \toprule
    \textbf{Failure mode} & \textbf{Governor} & \textbf{Edit} & \textbf{Fine–tune term(s)} \\
    \midrule
    Flip–flop (heading) & $H$-step commitment  & steer away from & $\lambda_1\,\text{churn}$ \\
    &window &flip  logits &\\
    Dithering/paralysis & tie-break dwell;  & boost value margin  & $\lambda_3\,\text{freeze\_ticks}$ \\
    &min-step  rule &on ties &\\
    Corridor thrash & commit past doorway; & sparsify local layer & $\lambda_2\,\text{goal\_switches}$ \\
    &margin  gate& &\\
    Hypervigilance & pause budget then & temper uncertainty  & CVaR / near-tie  \\
    &commit  &response &penalties\\
    Detour bloat & bound detour & steer cost head & $\lambda_4\,\text{detour\_inflation}$ \\
    &  inflation& &\\
    Energy overuse & enforce rest on low  & steer away from & $\lambda_5\,\text{energy/thermal}$ \\
    & margin&costly  actions &\\
    Belief incoherence & confidence-weighted & n/a & consistency  \\
    &fusion   + veto& &regulariser\\
    Metric mismatch & calibrated costs/ & n/a & mismatch \\
    &region  tagging& &penalties \\
    \bottomrule
  \end{tabular}
 \caption{Failure modes and preferred repair levers. “Governor” = symbolic GP rules; “Edit” = activation steering/sparsity; “Fine–tune” = neurosis-aware loss on counterexamples.}
  \label{tab:failure-repair}
\end{table}

\subsection{Verification and runtime guardrails}
\label{subsec:verification-guardrails}

Repairs are accepted only if they: (i) monotonically improve neurosis metrics across the full destructive suite; (ii) show no regression on nominal tasks; and (iii) pass \emph{causal ablations} (patch off $\Rightarrow$ failure returns; patch on $\Rightarrow$ failure vanishes). We then keep thin \emph{runtime shields}: state/action filters for hard constraints, commitment windows gated by threat probes, and confidence gating to a conservative subpolicy under high ensemble disagreement. Finally, we schedule periodic ``reality checks'' (a small destructive batch) to catch drift and trigger small, automatic fine-tunes when needed.

\subsection{Is this just a GP band-aid?}
\label{subsec:not-a-bandaid}

No. The GP governor is the \emph{first} step: it provides immediate, legible relief and a normative target. The full loop is: diagnose $\to$ localize $\to$ govern (stabilize) $\to$ steer/edit $\to$ distill $\to$ fine-tune $\to$ verify $\to$ guardrail. The end state is a policy that behaves well \emph{by itself}, with a thin safety net—not a permanent crutch.

\section{Suggested Loop for Ongoing Repair}
\label{sec:ongoing-repair}

Building on the diagnostic bank from Section~\ref{sec:gp-psychoanalysis} and the three–track repair actions (govern / edit / fine–tune) in Section~\ref{subsec:three-track-repair}, we operationalise a cyclical process for long-term maintenance. The GP governor is not a one-time intervention but a reusable scaffold for adaptive repair. As the system continues to learn and encounter novel contexts, new neuroses may emerge—requiring fresh diagnosis and stabilisation. We therefore propose a repair loop that integrates symbolic governance, distillation, and fine-tuning into the system’s long-term maintenance strategy:

\begin{enumerate}
\item \textbf{Monitor for drift and emergent neuroses.} Use probe metrics and destructive suites to detect behavioural degradation or new failure modes.
\item \textbf{Update the counterexample bank.} Curate new minimal triggers that elicit the observed pathology.
\item \textbf{Re-synthesise the GP governor.} Apply multi-objective Genetic Programming to generate a symbolic controller that reduces updated neurosis metrics while preserving task competence.
\item \textbf{Deploy and rehearse.} Interpose the governor to stabilise behaviour under the new triggers, using its decisions as normative demonstrations.
\item \textbf{Distil into policy.} Clone the governor’s decisions into the policy via behavioural cloning or DAgger-style rollouts.
\item \textbf{Fine-tune with neurosis-aware loss.} Retrain on the updated counterexample bank, penalising pathological dynamics and reinforcing the distilled behaviours.
\item \textbf{Verify and re-guard.} Re-run causal ablations, confirm no regression on nominal tasks, and update runtime guardrails accordingly.
\end{enumerate}

This loop can be scheduled periodically, triggered by anomaly detection, or run continuously. The symbolic governor serves as a flexible, auditable scaffold for iterative repair—providing immediate relief and a normative target for internalisation. Over time, the policy becomes increasingly self-sufficient, with the governor receding to a thin safety net.

\section{Repairing the machine: analogy to a successful human psychoanalysis}
\label{sec:psychoanalytic-analogy}

As a high-level mnemonic for Section~\ref{sec:ongoing-repair}, clinical psychoanalysis proceeds through exposure and naming of triggers, interpretation of underlying conflicts, rehearsal of healthier responses in safe settings, internalisation of change, and relapse prevention. Our loop mirrors this arc. Destructive testing provides \emph{exposure and naming} (minimal counterexamples with measured symptoms). Probe-driven localisation supplies \emph{interpretation} (which internal representations misvalue threat, energy, or commitment). The GP governor offers \emph{rehearsal}—simple rational constraints that demonstrate healthier behaviour under the same triggers. Distillation and targeted fine-tuning achieve \emph{internalisation}, transferring those behaviours into the policy so the governor can recede. Finally, shields and scheduled reality checks furnish \emph{relapse prevention}. In this sense, the system is not merely patched; it is \emph{treated}: symptoms are stabilised, causes are understood, habits are retrained, and safeguards remain to prevent recurrence.


\begin{thebibliography}{99}
\bibitem{Sutton2019}Sutton, R. S. (2019). The bitter lesson. \url{http://www.incompleteideas.net/IncIdeas/BitterLesson.html}
\bibitem{Asimov1950}Asimov, I. (1950). I, Robot. New York: Gnome Press. (Original story “Runaround” published in Astounding Science Fiction, 1942)
\bibitem{Solms2021} Solms, M. (2021). The hidden spring: A journey to the source of consciousness. New York: W. W. Norton \& Company. ISBN: 9780393542028
\bibitem{Solms2012} Solms, M. \& Panksepp, J. (2012). The ``Id'' knows more than the ``Ego'' admits. \emph{Brain Sciences}, 2(2), 147–175. \url{https://pmc.ncbi.nlm.nih.gov/articles/PMC4061793/}
\bibitem{McGaugh2004} McGaugh, J. L. (2004). The amygdala modulates consolidation of memories. \emph{Annual Review of Neuroscience}, 27, 1–28. \url{https://pubmed.ncbi.nlm.nih.gov/15217324/}
\bibitem{McGaugh2015} McGaugh, J. L. (2015). Consolidating memories: The role of arousal. \emph{Frontiers in Psychology}, 6:1784. \url{https://pubmed.ncbi.nlm.nih.gov/25559113/}
\bibitem{Kaczkurkin2017} Kaczkurkin, A. N. et al. (2017). Neural substrates of overgeneralized conditioned fear in PTSD. \emph{American Journal of Psychiatry}, 174(2), 125–134. \url{https://psychiatryonline.org/doi/10.1176/appi.ajp.2016.15121549}
\bibitem{Lissek2014} Lissek, S. et al. (2014). Overgeneralization of conditioned fear as a pathological marker of anxiety. \emph{Biological Psychology}, 100, 1–12. 
\bibitem{Koza1994} Koza, J. R. (1994). \emph{Genetic Programming II: Automatic Discovery of Reusable Programs}. MIT Press. (Review: \url{https://direct.mit.edu/artl/article/1/4/439/2231/})
\bibitem{Koza1996} Koza, J. R. (1996). Use of Automatically Defined Functions and architecture-altering operations in GP. \emph{Pacific Symposium on Biocomputing}. \url{https://psb.stanford.edu/psb-online/proceedings/psb96/koza.pdf}
\bibitem{RobertsHowardKoza2001}
Roberts, S. C., Howard, D., \& Koza, J. R. (2001).
Evolving Modules in Genetic Programming by Subtree Encapsulation.
In: Miller, J., Tomassini, M., Lanzi, P. L., Ryan, C., Tettamanzi, A. G. B., \& Langdon, W. B. (eds)
\emph{Genetic Programming} (EuroGP 2001), LNCS 2038, pp.\ 160--175.
Springer, Berlin, Heidelberg. \url{https://doi.org/10.1007/3-540-45355-5_13}
\bibitem{Howard2003}
Howard, D. (2003).
Modularization by Multi-Run Frequency Driven Subtree Encapsulation.
In: Riolo, R., \& Worzel, B. (eds) \emph{Genetic Programming Theory and Practice},
Genetic Programming Series, vol 6, pp.\ 155--171.
Springer, Boston, MA. \url{https://doi.org/10.1007/978-1-4419-8983-3_10}
\bibitem{RussellNorvig2021} Russell, S. J., \& Norvig, P. (2021). \emph{Artificial Intelligence: A Modern Approach} (4th ed.). Pearson. 
\bibitem{Chebyshev1857} Chebyshev, P. L. (1857). Sur les meilleures approximations des fonctions par des polynômes. Journal de Mathématiques Pures et Appliquées.
\bibitem{Minkowski1896} Minkowski, H. (1896). Geometrie der Zahlen. (Collected/expanded in 1910. Lectures from the 1890s primary source for the Minkowski metric.)
\bibitem{DezaDeza2009} Deza, M. M. and Deza, E. D. (2009). Encyclopedia of Distances, Springer 1st Edition.
\bibitem{GonzalezWoods2018} Gonzalez, R. C., and Woods, R. E. (2018). Digital Image Processing, Pearson 4th Edition.
\bibitem{Moore1962} Moore, E. F. (1962). Machine Models of Self-Reproduction. \emph{Proceedings of Symposia in Applied Mathematics}. 14, 17-33, American Mathematical Society.
\bibitem{vonNeumann1966} von Neumann, J. (1966).  Theory of Self-Reproducing Automata. Editor: Arthur W. Burks. University of Illinois Press.
\bibitem{Krause1975}Krause, E. F. (1975). Taxicab Geometry. Addison–Wesley.
\bibitem{Ilachinski2001}Ilachinski, A. (2001). Cellular Automata: A Discrete Universe, World Scientific.
\end{thebibliography}
\end{document}